\pdfoutput=1
\documentclass[11pt]{article}

\usepackage[final]{emnlp}

\usepackage{times}
\usepackage{latexsym}

\usepackage[T1]{fontenc}

\usepackage[utf8]{inputenc}

\usepackage{microtype}

\usepackage{inconsolata}

\usepackage{graphicx}

\usepackage{booktabs}
\usepackage{colortbl}
\usepackage{tabularx}
\usepackage{array}
\usepackage{placeins}

\newcolumntype{Y}{>{\centering\arraybackslash}X}

\title{SciFigBench: A Benchmark for Scientific Method Figure Understanding}

\author{
  \textbf{Lijia Wu}\textsuperscript{1,2},
  \textbf{Zheng Li}\textsuperscript{3,*},
  \textbf{Yuxuan Liu}\textsuperscript{1,2},
  \\
  \bfseries
  \textbf{Jingwen Zhang}\textsuperscript{1,2},
  \textbf{Penghai Zhao}\textsuperscript{1,2},
  \textbf{Xiang Li}\textsuperscript{2,1,*}
  \\
  \textsuperscript{1}VCIP, CS \& AAIS, Nankai University \\
  \textsuperscript{2}NKIARI, Shenzhen Futian \quad
  \textsuperscript{3}Li Auto. \\
  \texttt{lijiawu@mail.nankai.edu.cn} \\
  \texttt{zhengli97@foxmail.com} \quad
  \texttt{xiang.li.implus@nankai.edu.cn}
}

\begin{document}
\maketitle
\begingroup
\renewcommand{\thefootnote}{\fnsymbol{footnote}}
\footnotetext[1]{Corresponding authors.}
\endgroup

\begin{abstract}
Scientific method figures often provide researchers with a clear visual entry point into a paper's methodology and core contributions, yet their dense layouts and complex information flows are difficult for Vision Language Models (VLMs) to interpret.
Existing benchmarks predominantly focus on coarse-grained tasks, such as context-based retrieval or holistic comprehension, overlooking the fine-grained content and inter-module relations within complex methodological diagrams.
To address these overlooked aspects, we introduce \textbf{Sci}entific \textbf{Fig}ure \textbf{Bench}mark (\textbf{SciFigBench}), a benchmark for evaluating core capabilities required for scientific method figure understanding.
Specifically, it targets three capabilities: dense content perception, visual relation comprehension, and information-flow tracing.
For this purpose, we curate 4{,}352 scientific method figures from real-world research papers and construct 40{,}670 QA pairs to operationalize these capabilities as measurable question-answering tasks.
All annotations are generated by dedicated pipelines and then manually verified.
We further propose \textbf{SciFigCue}, a training-free visual prompting module that externalizes arrow-induced dependencies as visual relational evidence for method-figure reasoning.
Experiments across open-source and proprietary VLMs show that even strong models remain substantially limited in visual relation comprehension, while SciFigCue improves performance across several tasks without parameter updates.
The code and related resources are available at \url{https://github.com/LeNcyer/SciFigBench}.

\end{abstract}

\section{Introduction}
\label{sec:introduction}

As AI systems become increasingly involved in scientific research, recent work has explored their potential across research workflows~\citep{garikaparthi-etal-2025-mir,zhang-etal-2024-comprehensive-survey,naipv2,zhao-etal-2025-abgen,wang-etal-2024-scimon}.
To be useful in these settings, such systems must understand the methods described in papers, rather than only retrieve factual information.
This requires models to parse how papers organize procedural steps and communicate technical ideas.
In research papers, method figures are among the most compact forms of method communication, organizing model architectures, processing pipelines, and technical ideas into a single diagram.
Their semantics are not defined by individual visual objects, but by the interaction between text labels, visual modules, and arrow connections.
Thus, method figure understanding requires recovering both local content and directed relations, rather than merely recognizing elements in the image.

\begin{figure}[t]
\centering
\makebox[\columnwidth][c]{\includegraphics[height=0.30\textheight,trim=150pt 10pt 150pt 10pt,clip]{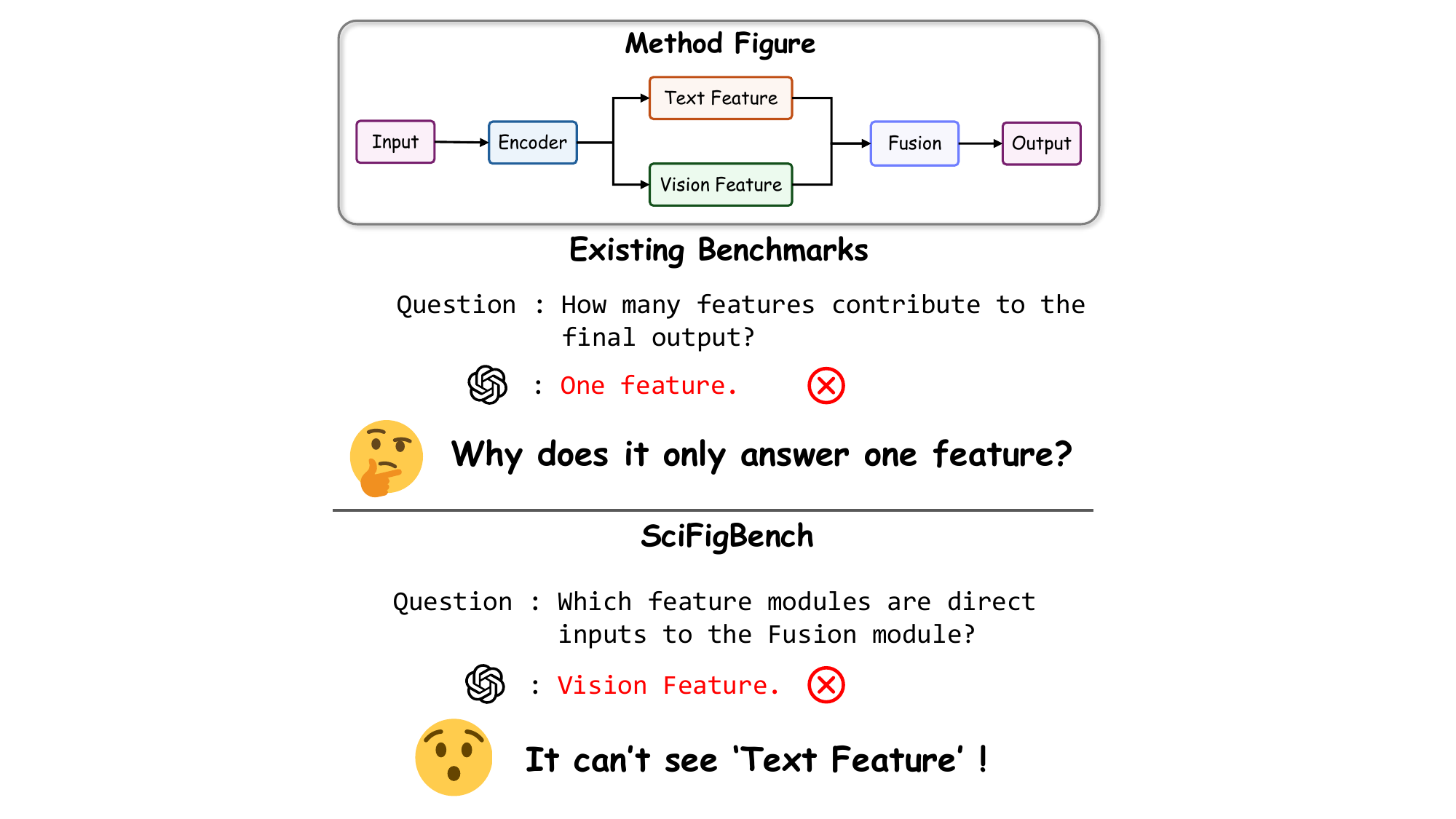}}
\caption{\textbf{Why diagnostic method-figure QA is needed.}
A coarse-grained question may expose an incorrect answer, but cannot distinguish whether the failure comes from missing a component label or misreading a structural connection.
SciFigBench instead asks fine-grained diagnostic questions to make these failure sources observable.}
\label{fig:benchmark-compare}
\end{figure}

This demand for fine-grained diagram understanding has motivated recent benchmarks on scientific diagrams and method figures in research papers~\citep{missqa,flowlearn,scicqa}.
Existing benchmarks cover several evaluation formats, including information-seeking QA over schematic diagrams, holistic figure-level comprehension, and structural reasoning over flowchart-style synthetic diagrams.
These works provide important testbeds and show that current VLMs still have gaps in scientific figure understanding.

However, final task scores indicate only whether a model answers correctly, offering little insight into the specific sources of failure in method figure understanding.
The same wrong answer may arise from different causes: a model may miss local text, misread an arrow connection, or fail to follow a multi-step information flow.
These failures correspond to different visual reasoning abilities, yet existing evaluations usually collapse them into a single aggregate score.

To diagnose such failures, we introduce \textbf{SciFigBench}, a \textbf{Sci}entific \textbf{Fig}ure \textbf{Bench}mark for evaluating VLMs' ability to understand real scientific method figures.
Specifically, SciFigBench is organized around three complementary evaluation perspectives: 
(1) \textbf{dense content perception}, assessing whether models can capture textual cues distributed across method components; 
(2) \textbf{visual relation comprehension}, measuring their ability to identify module connections, local connectivity patterns, and directional interactions; 
and (3) \textbf{information-flow tracing}, testing whether models can integrate textual cues with structural relations to trace input--output flows across method modules.
By separating these perspectives, SciFigBench links model failures to distinct visual reasoning abilities.

\begin{figure}[t]
\centering
\includegraphics[width=0.8\columnwidth,trim=250pt 20pt 250pt 20pt,clip]{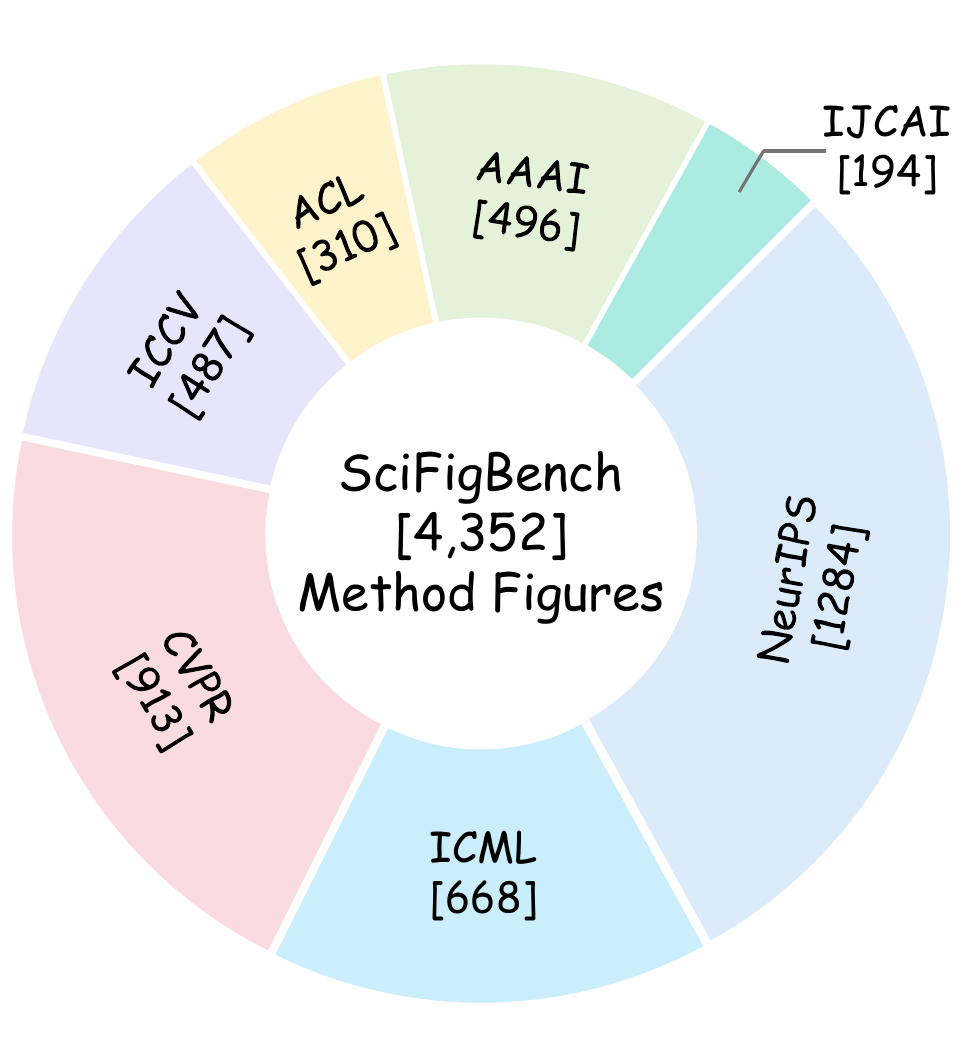}
\caption{\textbf{Source distribution of SciFigBench method figures.}
SciFigBench contains 4{,}352 method figures from papers published at seven AI conferences in 2024--2025.}
\label{fig:paper-distribution}
\end{figure}

Building such a diagnostic benchmark requires structured annotations of figure elements and their directed relations, which are rarely available from real method figures.
Therefore, we introduce an automated annotation pipeline that generates initial annotations for figure text, visual components, and their connections.
All generated candidate annotations are then manually verified to ensure the quality of the final evaluation data.
This pipeline enables the construction of SciFigBench, comprising 4{,}352 scientific method figures curated from 2024--2025 AI conference papers and 40{,}670 QA pairs across the three diagnostic dimensions.
We use AI method figures as an initial testbed because text-dense, free-form layouts and directed information flows closely align with the figure-internal capabilities targeted by SciFigBench.
Figure~\ref{fig:paper-distribution} summarizes the distribution of these method figures across seven venues.

Experiments on SciFigBench reveal that strong proprietary and open-source VLMs still struggle with fine-grained relation understanding and information-flow tracing in scientific method figures.
These failures point to arrow-based connection cues as a key difficulty: a model must detect arrows, infer their directions, and compose the induced connections to follow relations across method components.

This diagnosis points to a simple input-level enhancement through visual prompting~\citep{shtedritski2023does}: making the missing connection cues explicit.
Therefore, we introduce \textbf{SciFigCue}, a training-free visual prompting module for scientific method figures.
Specifically, SciFigCue renders arrow-based connection cues in two forms.
It overlays arrow cues on the original figure to preserve visual context, and provides an auxiliary arrow-segmentation view to make arrow locations and head--tail directions explicit.
As a training-free input preprocessing module, SciFigCue can be seamlessly integrated into existing VLMs without additional training cost or model modification.
Extensive experiments show that this simple plug-in improves multiple connection-dependent tasks, suggesting that explicit connection cues strengthen structured reasoning over method figures.

Our contributions are as follows:
\begin{itemize}
    \item We introduce \textbf{SciFigBench}, a diagnostic benchmark for evaluating VLMs' figure-internal understanding of real scientific method figures.

    \item We construct a human-verified dataset with 4{,}352 real scientific method figures and 40{,}670 QA pairs.

    \item We propose \textbf{SciFigCue}, a training-free visual prompting module that makes arrow-based connection cues explicit for VLM reasoning.

    \item Experiments show that current VLMs struggle to compose arrow-induced connections, while SciFigCue's explicit connection cues help alleviate this bottleneck.
\end{itemize}

\section{Related Work}
\label{sec:related_work}

\subsection{Structured Visual Understanding Benchmarks}

Structured visual understanding benchmarks have expanded from charts~\citep{chartqa,chartmind} to flowcharts and scientific figures.
FlowchartQA~\citep{flowchartqa}, FlowVQA~\citep{flowvqa}, FlowLearn~\citep{flowlearn}, and FlowExplainBench~\citep{flowexplainbench} focus on reasoning over procedural structures, control flows, and graph-like relations.
FlowLearn is closest to our setting, but its fine-grained topology analysis is mainly built on simulated flowcharts, while its real scientific-figure subset is evaluated primarily through OCR, true/false QA, and caption-based description.
MISS-QA~\citep{missqa} and SciCQA~\citep{scicqa} extend evaluation to schematic diagrams and scientific charts in research papers, but their objectives remain different from diagnosing the internal structure of method figures.
Overall, existing chart, flowchart, and scientific-figure benchmarks provide important testbeds, but they do not directly diagnose localized text, component relations, and information flow in real scientific method figures.
SciFigBench fills this gap by directly annotating localized text fields and directed component relations in real method figures and constructing diagnostic tasks around them.
\subsection{Visual Prompting for Multimodal Reasoning}

As VLMs~\cite{radford2021learning,liu2023visual,zhai2023sigmoid,liu2024improved,li2024llava} develop stronger visual reasoning and perception capabilities~\citep{zhang2026crystal,zhang2025unichange,li2026wowseg}, visual prompting~\cite{zhou2022learning,jia2022visual,shtedritski2023does,li2024promptkd,li2025advancing} has been used to improve grounding and region understanding with cues such as color, shape, position, discrete marks, boxes, points, and arrows~\citep{shtedritski2023does,setofmarkprompting,vipllava}.
These methods mainly target generic object grounding, region understanding, or broad visual enhancement, rather than directed connection reasoning in method figures.
SciFigCue targets this structured setting by reusing the automatic annotation pipeline behind SciFigBench to render arrow locations, head--tail directions, and query-relevant target regions as visual prompts for relation understanding and information-flow tracing.

\section{SciFigBench}
\label{sec:scifigbench}

We introduce SciFigBench, a diagnostic benchmark for scientific method figure understanding.
It decomposes this capability into a set of QA tasks designed to reveal where model reasoning fails.
The detailed construction of SciFigBench is described below.

\subsection{Dataset Construction}
\label{sec:scifigbench:dataset}

\begin{figure*}[t]
\centering
\includegraphics[width=\textwidth,trim=105pt 25pt 105pt 25pt,clip]{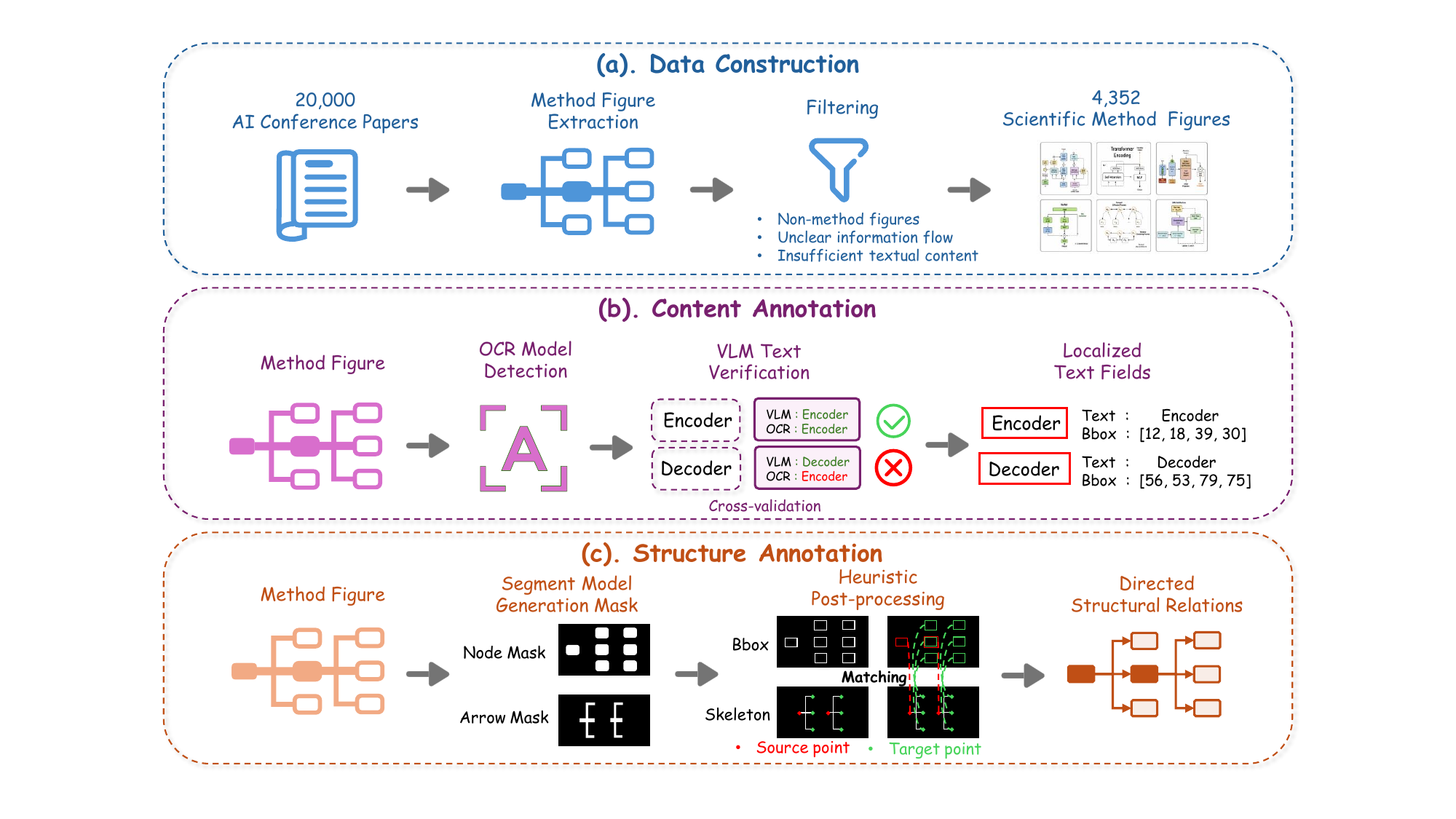}
\caption{\textbf{Overview of the SciFigBench construction pipeline.}
\textbf{(a)} Starting from 20{,}000 AI conference papers, we extract candidate method figures and retain 4{,}352 scientific method figures after filtering.
\textbf{(b)} Content annotation combines OCR detection with VLM-based verification to produce localized text fields.
\textbf{(c)} Structural annotation extracts node boxes and head/tail-labeled arrow endpoints through segmentation and post-processing to build directed structural relations.
Collectively, these annotations enable the construction of the diagnostic QA tasks in SciFigBench.}
\label{fig:overview}
\end{figure*}

Figure~\ref{fig:overview} gives an overview of the SciFigBench construction pipeline, which consists of figure collection, content annotation, and structural annotation.
Both annotation types are produced by dedicated automated pipelines and manually verified before being used to construct diagnostic QA examples.
This process provides SciFigBench with localized textual content and directed structural relations for real method figures.

\paragraph{Data collection.}
We collect figures from AI conference papers published in 2024--2025, where method figures often present complex methods in compact visual form.
We start from 20,000 paper PDFs, extract candidate primary method figures from figure captions, and discard figures that are not method-centric, lack clear information flow, or contain insufficient textual content.
After filtering, 4{,}352 real scientific method figures constitute our evaluation dataset.

\paragraph{Content annotation.}
The content annotation records localized text fields in each figure, where each field is represented by a text string and its bounding box.
We construct this annotation through OCR--VLM cross-validation because method figures often contain small, densely packed text.
We first use an OCR model~\citep{paddleocr} to detect text regions and produce OCR hypotheses; a verifier VLM\citep{qwen3vl} then reads each cropped region and selects the more reliable candidate.
The content annotation provides text-level supervision for dense content perception.
These annotations also support diagnostic tasks that require integrating textual evidence with figure structure.

\paragraph{Structural annotation.}
The structural annotation captures visual modules, arrow endpoints, and directed connections in each method figure.
Each visual module is localized by a node box, and each arrow is represented by a tail endpoint and a head endpoint.
Directed connections between nodes are defined by the correspondence between arrow endpoints and node boxes.
To construct this annotation, we first apply a segmentation model~\citep{sam3} to generate node and arrow masks, and convert node masks into boxes.
The arrow masks are then skeletonized to recover endpoints, and each endpoint neighborhood is cropped for a VLM to determine whether it is the arrow head or tail.
Finally, endpoints are associated with node boxes via nearest-boundary matching, yielding candidate directed connections from tail-matched nodes to head-matched nodes.
The structural annotation provides structural supervision for visual relation comprehension and information-flow tracing tasks.

\begin{figure}[t]
\centering
\includegraphics[width=0.9\columnwidth,trim=275pt 65pt 275pt 65pt,clip]{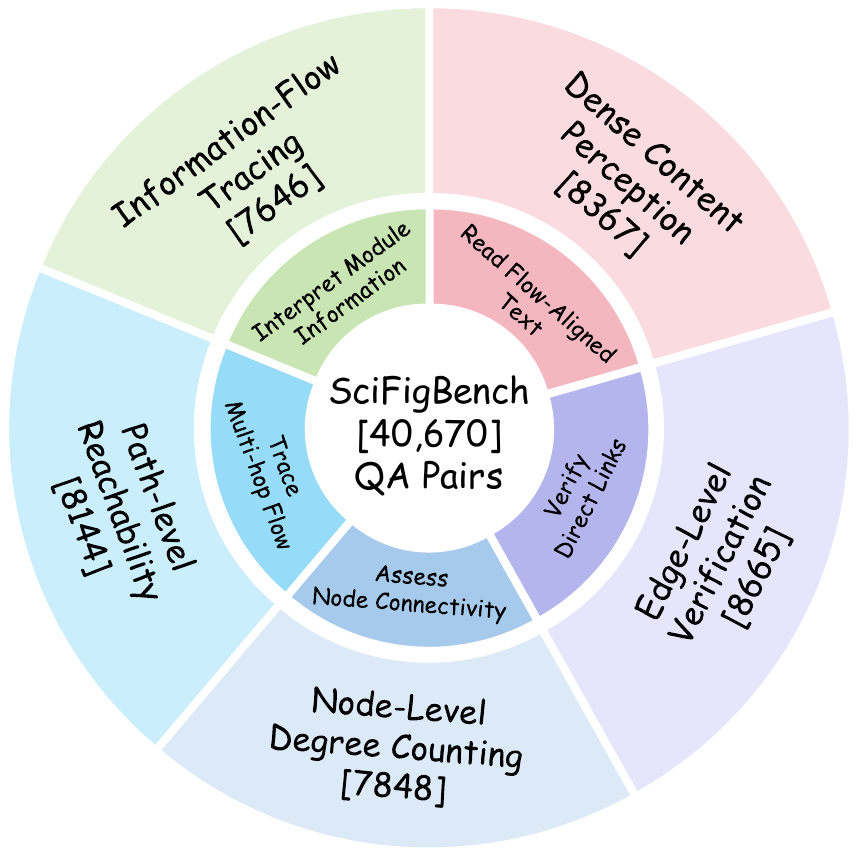}
\caption{\textbf{QA distribution in SciFigBench.}
The 40{,}670 QA pairs are distributed across five diagnostic tasks grouped into three evaluation dimensions: dense content perception, visual relation comprehension, and information-flow tracing.}
\label{fig:qa-distribution}
\end{figure}

All automatically generated content and structural annotations are manually verified by 20 annotators, each receiving over three hours of calibration training before verification.
Human verification shows that the automated annotation pipeline achieves an overall F1 of 82.21\% across the two annotation types.
Detailed verification results are provided in Appendix~\ref{sec:appendix:annotation_verification}.
This automatic-first workflow focuses human effort on candidate verification rather than annotation from scratch, reducing annotation cost.
The verified annotations are then used to construct 40{,}670 QA pairs across five diagnostic QA tasks.
Figure~\ref{fig:qa-distribution} shows the distribution of QA pairs across these five tasks.
Representative annotation examples are provided in Appendix~\ref{sec:appendix:annotation_examples}.

\begin{figure*}[t]
\centering
\includegraphics[width=\textwidth,trim=10pt 180pt 10pt 180pt, clip]{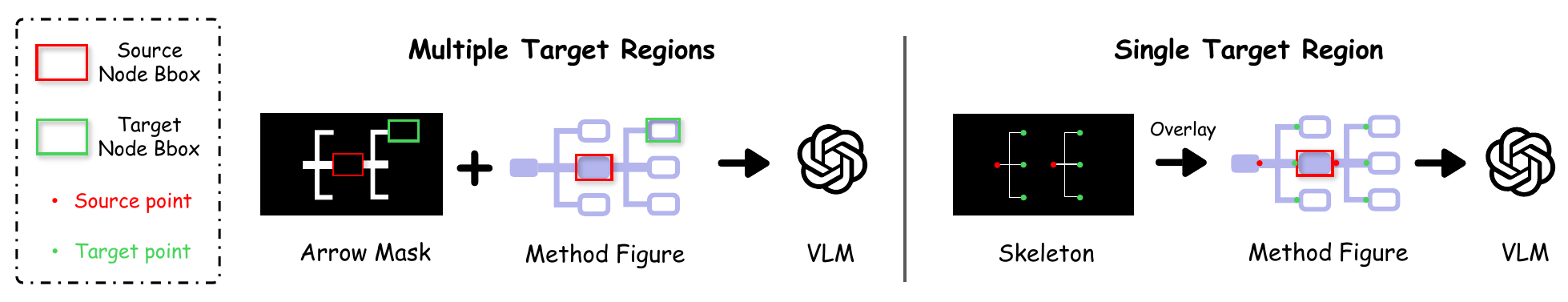}
\caption{\textbf{Two visual-prompt formats in SciFigCue.}
Visual cues are extracted by the automated parsing pipeline used in SciFigBench construction and rendered in two formats: an auxiliary arrow-mask view for multi-target queries and an original-figure overlay for single-target queries.}
\label{fig:scifigcue}
\end{figure*}

\subsection{Evaluation Tasks}
\label{sec:scifigbench:tasks}

SciFigBench decomposes method-figure understanding into three complementary evaluation perspectives for localizing different capability bottlenecks.
These perspectives cover dense content perception, visual relation comprehension, and information-flow tracing.
Detailed case construction, case statistics, and task prompts are provided in Appendix~\ref{sec:appendix:case_stats} and Appendix~\ref{sec:appendix:prompts}.

\subsubsection{Dense Content Perception}
\label{sec:scifigbench:dense_content}

Dense content perception evaluates whether VLMs can capture textual cues distributed across method components along an information-flow path.
This task reflects how human researchers read method figures when following a method flow: they read component text in context rather than recognize isolated text regions.

\paragraph{Task formulation.}
Given a method figure, we first select a flow path $P$ from its structural annotation, and then choose an ordered node subsequence $V_{\mathrm{sub}}=(v_1, v_2, \ldots, v_K)$ of $K$ text-bearing nodes from $P$.
For each selected node $v_i$, the aligned text field is denoted as $(t_i,b_i)$, where $t_i$ is the text string and $b_i$ is its region box.
The task input is the figure with region boxes $(b_1, b_2, \ldots, b_K)$ highlighted by a fixed color sequence following the node order in $V_{\mathrm{sub}}$.
The model must output the corresponding text list $[t_1, t_2, \ldots, t_K]$ in color order.
This task requires the model to localize distributed regions, read local text, and preserve the ordering induced by the information flow.

\paragraph{Sampling and metric.}
For each figure, we sample from the longest flow path to cover a more complete method flow.
The path length is limited to $K \le 6$.
When the number of candidate text-bearing nodes exceeds six, we sample uniformly across text-length buckets to cover different reading difficulties.
This task yields 8{,}367 QA pairs with broad coverage across different values of $K$.
The primary metric is position-wise ANLS~\citep{anls} averaged across queried positions.

\subsubsection{Visual Relation Comprehension}
\label{sec:scifigbench:relation}

Visual relation comprehension evaluates whether VLMs can understand arrow-induced structural relations between method components.
A central operation in reading method figures is to follow arrows and trace how information moves between components.
To evaluate this ability at different levels of structural complexity, we construct three subtasks from the structural annotation: edge-level verification, node-level degree counting, and path-level reachability.

\paragraph{Task formulation.}
For each method figure, the structural annotation provides a directed graph $G=(V,E)$, where each node $v \in V$ is localized by a corresponding region box and $E$ denotes arrow-induced directed connections.
Visual relation queries highlight the region boxes of one or two target nodes on the original figure.
The model then reasons about structural relations in $G$ based on the visible arrows.
The three subtasks evaluate the model's ability to determine whether a candidate edge $e=(v_s,v_t)$ belongs to $E$, predict the in-/out-degree of a target node $v$, and determine whether $v_t$ is reachable from $v_s$.
The model outputs either a binary decision or a non-negative integer, depending on the query type.
All three subtasks are evaluated using accuracy.

\paragraph{Edge-Level Verification.}
The most local form of relation understanding is determining whether a directed connection exists between two components.
Edge-Level Verification evaluates the model's ability to recognize directed relations between two method components.
Given a figure where the region boxes of a candidate source node $v_s$ and target node $v_t$ are highlighted in red and blue, respectively, the model must decide whether $(v_s,v_t) \in E$.
We balance positive and negative examples and construct hard negatives, yielding 8{,}665 QA pairs.

\paragraph{Node-Level Degree Counting.}
Beyond a single edge, the local role of a component also depends on how many connections enter or leave it.
Node-Level Degree Counting evaluates the model's ability to aggregate local connectivity around a target component.
Given a figure where the region box of a target node $v$ is highlighted and a query direction $d \in \{\texttt{in},\texttt{out}\}$ is specified, the model must output the number of arrows incident to the target node in that direction.
We bucket examples by the target degree to cover diverse local connectivity patterns, yielding 7{,}848 QA pairs.

\paragraph{Path-Level Reachability.}
At a higher level, relation understanding requires composing multiple directed connections to determine longer-range dependencies between components.
Path-Level Reachability primarily evaluates the model's ability to reason along directed connections beyond direct adjacency.
Given a figure where the region boxes of a source node $v_s$ and a target node $v_t$ are highlighted in red and blue, respectively, the model must determine whether a directed path exists from $v_s$ to $v_t$.
Positive and negative examples are carefully designed to cover diverse reachability patterns, resulting in 8{,}144 QA pairs.

\subsubsection{Information-Flow Tracing}
\label{sec:scifigbench:flow_tracing}

Information-flow tracing further requires models to jointly reason over textual evidence and directed structure.
Human researchers often read method figures by anchoring on a named component and tracing information flow along its incoming and outgoing connections to understand the component's role in the method flow.
To evaluate this ability, we design a neighbor-enumeration QA task that requires named-node localization, directional tracing, and neighbor text recognition without visual box prompts.

\paragraph{Task formulation.}
Given a method figure, a target node specified by its exact textual label \texttt{<TARGET>}, and a query direction $d \in \{\texttt{in}, \texttt{out}\}$, the model must return the textual contents of all direct neighboring nodes in the queried direction.
For $d=\texttt{in}$, the answer contains the text from all text-bearing nodes that directly point to the target; for $d=\texttt{out}$, it contains the text from all text-bearing nodes directly pointed to by the target.
The output is an unordered string list $[t_1, t_2, \ldots, t_n]$.
The target label must be unique within the figure; this constraint focuses the task on information-flow tracing after named-node localization.

\paragraph{Sampling and metric.}
To construct valid queries, we sample targets only from text-bearing nodes with unique textual labels and create paired in/out queries for each target.
During sampling, we match the distribution of candidate-neighbor counts across the in/out directions.
The proportion of empty answers is also controlled to reduce potential shortcuts arising from directional bias or empty-answer frequency.
The resulting task comprises 7{,}646 QA pairs.
Set-based ANLS$^{\star}$~\citep{anls-star} is used as the evaluation metric to account for the unordered nature of the answers.

\section{SciFigCue}
\label{sec:scifigcue}

To make fine-grained connection cues in method figures explicit to VLMs, we introduce \textbf{SciFigCue}, a training-free input preprocessing module.

\begin{table*}[t]
\centering
\tiny
\newcommand{\taskhead}[2]{%
  \begin{tabular}[c]{@{}c@{}}
    #1\\
    #2~
  \end{tabular}
}
\newcommand{\gainfont}{\fontsize{4.2}{4.6}\selectfont}
\newcommand{\posgain}[2]{#1{\gainfont\textcolor{green!50!black}{\,(#2)}}}
\newcommand{\neggain}[2]{#1{\gainfont\textcolor{red}{\,(#2)}}}
\setlength{\tabcolsep}{4pt}
\resizebox{\textwidth}{!}{%
\begin{tabular}{l c c c c c c}
\toprule
\textbf{Model} & \textbf{Size} & \textbf{\taskhead{Dense Content}{Perception}} & \textbf{\taskhead{Edge-Level}{Verification}} & \textbf{\taskhead{Node-Level}{Degree Counting}} & \textbf{\taskhead{Path-Level}{Reachability}} & \textbf{\taskhead{Information-Flow}{Tracing}} \\
\midrule
\textsc{RandomChoice} & -- & -- & 50.0 & -- & 50.0 & -- \\
\midrule
\rowcolor{gray!10}
\multicolumn{7}{l}{\emph{Closed-source models}} \\
Gemini-2.5-Flash  & -- & 72.7 & 66.0 & 42.3 & \underline{55.3} & 42.4 \\
Gemini-3.1-Flash-Lite & -- & 71.2 & \underline{67.3} & \textbf{55.8} & 54.1 & \textbf{46.1} \\
GPT-5-mini        & -- & 51.3 & 49.8 & 24.8 & 49.3 & 26.0 \\
GPT-5.5           & -- & \underline{76.8} & 55.8 & 34.7 & 51.8 & 38.6 \\
\midrule
\rowcolor{gray!10}
\multicolumn{7}{l}{\emph{Open-source models}} \\
InternVL3.5-2B~\citep{internvl35}   & 2B & 32.5 & 36.4 & 31.0 & 46.4 & 17.6 \\
InternVL3.5-8B~\citep{internvl35}    & 8B & 45.9 & 44.7 & 35.8 & 41.5 & 27.2 \\
InternVL3.5-14B~\citep{internvl35}   & 14B & 53.7 & 47.7 & 35.7 & 46.3 & 26.1 \\
InternVL3.5-38B~\citep{internvl35}   & 38B & 52.2 & 47.8 & 40.9 & 45.4 & 26.4 \\
LLaVA-OV-1.5-4B~\citep{llavaonevision}  & 4B & 46.0 & 44.0 & 33.0 & 41.8 & 24.6 \\
LLaVA-OV-1.5-8B~\citep{llavaonevision}  & 8B & 54.8 & 45.4 & 34.5 & 40.2 & 24.5 \\
Qwen3-VL-4B~\citep{qwen3vl}       & 4B & 51.6 & 48.6 & 38.9 & 44.6 & 27.5 \\
Qwen3-VL-8B~\citep{qwen3vl}       & 8B & 55.1 & 48.9 & 39.3 & 46.1 & 31.3 \\
Qwen3-VL-32B~\citep{qwen3vl}      & 32B & 62.6 & 54.0 & 46.4 & 49.4 & 35.4 \\
Qwen3.5-VL-2B    & 2B & 19.8 & 46.2 & 22.0 & 49.2 & 16.3 \\
Qwen3.5-VL-4B    & 4B & 45.1 & 48.8 & 37.5 & 46.1 & 26.3 \\
Qwen3.5-VL-9B    & 9B & 61.2 & 50.8 & 41.4 & 46.3 & 32.8 \\
Qwen3.5-VL-27B   & 27B & 70.9 & 59.2 & 50.9 & 50.2 & \underline{43.6} \\
\midrule
\rowcolor{gray!10}
\multicolumn{7}{l}{\emph{Ours}} \\
\rowcolor{blue!4}
SciFigCue (Qwen3.5-VL-9B)  & 9B & \posgain{73.0}{+11.8} & \posgain{56.6}{+5.8} & \posgain{44.1}{+2.7} & \posgain{52.0}{+5.7} & \posgain{36.0}{+3.2} \\
\rowcolor{blue!8}
SciFigCue (Qwen3.5-VL-27B) & 27B & \posgain{\textbf{82.4}}{+11.5} & \posgain{\textbf{70.1}}{+10.9} & \posgain{\underline{51.1}}{+0.2} & \posgain{\textbf{56.0}}{+5.8} & \neggain{42.1}{-1.5} \\
\bottomrule
\end{tabular}
}%
\caption{\textbf{Main results on SciFigBench.}
All values are percentages; higher is better.
Best and second-best results in each column are shown in \textbf{bold} and \underline{underlined}, respectively.
Colored parentheses in SciFigCue rows indicate absolute changes over the corresponding base model.
Overall, current VLMs remain limited in method figure understanding, while SciFigCue improves performance by making connection cues explicit.}
\label{tab:main_results}
\end{table*}

\paragraph{Formulation.}
Given a scientific method figure $I$ and a QA query $q$, a standard VLM directly predicts an answer $a$ from $(I,q)$.
SciFigCue instead transforms $I$ into an enhanced visual input $\tilde{I}$ by explicitly marking arrow locations, head--tail directions, and query-relevant target regions.
The VLM then predicts the answer from $(\tilde{I},q)$.

\paragraph{Automatic Cue Extraction.}
SciFigCue reuses the automated parsing pipeline from SciFigBench construction at test time to extract intermediate parsing results for visual prompting.
These intermediate results include text or node regions for grounding query targets, arrow masks for isolating connection structure, and head--tail endpoints for marking direction.

\paragraph{Task-Adaptive Visual Prompting.}
SciFigCue uses two visual-prompt formats depending on whether the query is centered on a single target node or involves multiple target regions.
Figure~\ref{fig:scifigcue} illustrates the overall pipeline for these two prompting formats.
For multi-target queries, SciFigCue provides the original figure together with an auxiliary arrow-mask view.
The auxiliary view uses the arrow mask extracted by the segmentation model~\citep{sam3} as its background and overlays the target regions associated with the current query.
This representation reduces visual clutter and makes the connections between target regions more explicit.
This format is used for dense content perception, edge-level verification, and path-level reachability.
For single-target queries, SciFigCue overlays visually distinct arrow endpoint cues directly on the original figure, making incoming and outgoing connections around the target node explicit.
This format is used for node-level degree counting and information-flow tracing.
Additionally, for text-specified targets in information-flow tracing, SciFigCue matches the target name using the content annotation and highlights the corresponding region.
SciFigCue changes only the input representation, leaving the answer space, evaluation protocol, and model parameters unchanged.

\section{Experiments}
\label{sec:experiments}

\subsection{Experimental Setup}
\label{sec:experiments:setup}

We use SciFigBench to evaluate current VLMs on scientific method-figure understanding and to test whether explicit connection cues improve method-figure reasoning.
The SciFigBench evaluation comprises 40{,}670 QA pairs and follows the task inputs and output formats defined in Section~\ref{sec:scifigbench}.
In subsequent tables, we abbreviate the five evaluation tasks as DCP, Edge, Degree, Reach., and IFT.
We report position-wise ANLS for DCP, set-based ANLS$^{\star}$ for IFT, and accuracy for Edge, Degree, and Reachability.
The evaluated model set includes open-source families such as InternVL3.5, LLaVA-OV-1.5, Qwen3.5-VL, and Qwen3-VL, together with proprietary models from Gemini and GPT.
Open-source models are evaluated on the full test set, while proprietary models are evaluated on 2{,}000 QA pairs sampled proportionally by task due to inference costs.
Evaluation of open-source models on the same 2{,}000-QA subset shows only minor performance changes and leaves the overall conclusions unchanged.
Detailed subset results are reported in Appendix~\ref{sec:appendix:subset_evaluation}.

\subsection{Main Results}
\label{sec:experiments:main_results}

\paragraph{Overall performance.}

Table~\ref{tab:main_results} shows that SciFigBench remains challenging for current VLMs in both textual perception and structural reasoning.
Among models evaluated without SciFigCue, the best scores are 76.8 on Dense Content Perception and 46.1 on Information-Flow Tracing, indicating substantial room for improvement in dense text perception and text-structure integration.
Structural reasoning is even more challenging, with chance accuracy at 50.0 on both Edge-Level Verification and Path-Level Reachability.
Many strong models remain near chance on Path-Level Reachability, such as Qwen3.5-VL-27B at 50.2 and Qwen3-VL-32B at 49.4.
Node-Level Degree Counting is similarly difficult, with the best result reaching only 55.8.
Overall, current VLMs are limited not only in text perception but also in reliably aggregating and composing directed relations in method figures.
Such structural reasoning is straightforward for human readers, while current VLMs still struggle to perform it reliably.

\paragraph{Scaling behavior.}

Model scaling improves text perception more consistently, but does not reliably improve structural reasoning.
Across open-source families, larger models tend to perform better on Dense Content Perception and Information-Flow Tracing. Scaling Qwen3.5-VL from 2B to 27B yields gains of 51.1 and 27.3 points on these two tasks, respectively.
In contrast, this trend is much weaker for path-level structural reasoning. Across the same scale range, Path-Level Reachability improves by only 1.0 point.
This suggests that insufficient model capacity is not the only bottleneck in method-figure understanding. Current models still struggle to jointly reason over arrows, directions, and component layouts.

\paragraph{SciFigCue gains.}

The gains from SciFigCue further indicate that this bottleneck is not solely a matter of model scale alone.
Without updating model parameters, SciFigCue improves multiple structure-dependent tasks by making arrow locations, directions, and target regions explicit.
These gains suggest that current VLMs struggle to reliably extract and exploit fine-grained connection cues from raw method figures.
Therefore, the key challenge extends beyond reading figure text to translating arrow-induced visual structure into reliable relational reasoning.

\subsection{Diagnostic Analyses}
\label{sec:experiments:diagnostic_analyses}

\paragraph{Automatic parsing and symbolic reasoning.}

\begin{table}[t]
\centering
\small
\setlength{\tabcolsep}{2.5pt}
\begin{tabular}{@{}lccccc@{}}
\toprule
Method & DCP & Edge & Degree & Reach. & IFT \\
\midrule
Qwen3.5-VL-27B & 70.9 & 59.2 & 50.9 & 50.2 & 43.6 \\
AutoParse+Symbolic & 96.3 & 77.1 & 57.0 & 65.2 & 48.4 \\
\midrule
Verified Oracle & 100.0 & 100.0 & 100.0 & 100.0 & 100.0 \\
\bottomrule
\end{tabular}

\caption{\textbf{Automatic parsing and symbolic reasoning.}
AutoParse+Symbolic relies only on outputs from the automated parsing pipeline used in SciFigBench construction, with query regions matched to parsed nodes by geometry.
Verified Oracle uses human-verified annotations.
Both VLMs and AutoParse+Symbolic remain substantially below the oracle.}

\label{tab:autoparse_symbolic}
\end{table}

To separate structural parsing from downstream graph reasoning, we construct an AutoParse+Symbolic baseline that performs deterministic reasoning over automatically recovered graph structure.
This baseline uses only pre-verification parsing outputs, with no access to human-verified annotations or QA answers.
As shown in Table~\ref{tab:autoparse_symbolic}, AutoParse+Symbolic substantially outperforms Qwen3.5-VL-27B, yet both remain well below the oracle on graph-dependent tasks.
The remaining gap between AutoParse+Symbolic and the oracle primarily stems from upstream structure-recovery errors. Missing, spurious, or reversed edges directly affect downstream reasoning tasks.
The oracle results indicate that downstream graph reasoning is relatively straightforward once accurate graph structure is available.
Combined with the VLM results, these findings identify reliable recovery and use of directed information-flow structure from raw figures as the primary bottleneck.

\paragraph{Fine-grained diagnostic analysis.}

\begin{table}[t]
\centering
\small
\setlength{\tabcolsep}{4pt}
\begin{tabular}{lccc}
\toprule
Model & $K{=}1$ & $K{\ge}2$ Avg. & Gap \\
\midrule
Qwen3.5-VL-9B          & 76.0 & 45.5 & 30.5 \\
+ SciFigCue             & 84.0 & 61.5 & 22.5 \\
Qwen3.5-VL-27B         & 75.9 & 65.8 & 10.1 \\
+ SciFigCue             & 85.2 & 79.7 & 5.5 \\
\bottomrule
\end{tabular}

\caption{\textbf{$K$-wise analysis of Dense Content Perception.}
$K{\ge}2$ Avg. averages $K=2,\ldots,6$, and Gap denotes its difference from $K=1$.
SciFigCue consistently narrows the performance gap between the two settings.}

\label{tab:k_wise_analysis}
\end{table}

To probe model behavior beyond aggregate scores, we stratify Dense Content Perception results by the number of queried regions $K$.
A clear gap emerges between single- and multi-region reading: the $K=1$ score exceeds the $K\ge2$ macro average by 30.5 points for 9B and 10.1 points for 27B.
SciFigCue narrows these gaps to 22.5 and 5.5 points for 9B and 27B, respectively.
These results further show that current VLMs struggle to extract and exploit arrow-induced information-flow cues from raw method figures.
By making these cues explicit, SciFigCue substantially strengthens ordered text perception across multiple regions.
Further fine-grained analyses for the remaining tasks are provided in Appendix~\ref{sec:appendix:diagnostic_analysis}.

\subsection{SciFigCue Analysis}
\label{sec:experiments:scifigcue_analysis}

\paragraph{Controlled cue ablation.}

\begin{table}[t]
\centering
\small
\setlength{\tabcolsep}{3.5pt}
\begin{tabular}{lccc}
\toprule
Setting & Edge & Degree & Reach. \\
\midrule
Raw                     & 50.8 & 41.4 & 46.3 \\
Box only                & 48.3 & 40.7 & 47.9 \\
+ Arrow location        & 53.8 & 37.8 & 48.7 \\
+ Randomized arrow      & 49.9 & 41.3 & 46.8 \\
SciFigCue               & \textbf{56.6} & \textbf{44.1} & \textbf{52.0} \\
\bottomrule
\end{tabular}

\caption{\textbf{Controlled cue ablation on Qwen3.5-VL-9B.}
Raw denotes the standard SciFigBench input.
Within each task, the prompted variants share the same rendering design and differ only in cue content.
SciFigCue includes both arrow location and direction.
For Degree, Box only strengthens the target highlight without adding arrow cues.
Full SciFigCue performs best.}

\label{tab:cue_ablation}
\end{table}

Controlled ablations are conducted on Qwen3.5-VL-9B across the Visual Relation Comprehension subtasks, with only the cue content varied and all other settings fixed.
As shown in Table~\ref{tab:cue_ablation}, full SciFigCue achieves the best performance on all three tasks.
The randomized arrow cue preserves the same visual-prompt format but breaks its correspondence with the true arrow structure, reducing performance to near the Raw baseline.
This indicates that the gains arise from semantically correct arrow information rather than additional visual salience or arbitrary markers.
Arrow-location cues alone improve Edge and Reachability but degrade Degree, indicating task-dependent sensitivity to cue content.
This result suggests that edge and path level reasoning benefits from arrow layout, whereas in/out-degree counting relies more strongly on accurate direction.

\paragraph{Sensitivity to cue quality.}

\begin{table}[t]
\centering
\small
\setlength{\tabcolsep}{4pt}
\begin{tabular}{llcc}
\toprule
Model & Cue & Degree & IFT \\
\midrule
Qwen3.5-VL-9B  & Auto & 44.1 & 36.0 \\
Qwen3.5-VL-9B  & GT   & 46.1 & 36.9 \\
Qwen3.5-VL-27B & Auto & 51.1 & 42.1 \\
Qwen3.5-VL-27B & GT   & 52.7 & 43.6 \\
\bottomrule
\end{tabular}

\caption{\textbf{Sensitivity to arrow cue quality.}
Auto uses automatically parsed cues, while GT uses human-verified cues.
GT cues further improve SciFigCue performance, eliminating the IFT degradation observed with Auto cues on Qwen3.5-VL-27B.}

\label{tab:arrow_quality}
\end{table}

We further evaluate a ground-truth (GT) variant that replaces automatically parsed arrow locations and head--tail labels with human-verified cues while keeping the visual-prompt format fixed.
This setting isolates cue quality from the prompt format and model parameters.
This evaluation focuses on Degree and IFT, which rely on local arrow direction and relation tracing along information flow, respectively.
Table~\ref{tab:arrow_quality} shows that GT cues consistently improve performance across both VLMs and tasks.
This consistent improvement indicates that SciFigCue remains sensitive to cue fidelity, with more accurate arrow locations and directions providing stronger structural evidence for VLM reasoning.

\section{Conclusion}
\label{sec:conclusion}

This paper investigates whether current VLMs can reliably understand text, structural connections, and information flow in scientific method figures.
We introduce \textbf{SciFigBench}, a diagnostic benchmark that evaluates local content perception, structural relation understanding, and information-flow tracing in real method figures.
Experiments show that current VLMs not only miss textual evidence but also struggle to aggregate directed relations and trace multi-step dependencies, indicating that reliably recovering and using figure-internal connection structure remains a key bottleneck in method-figure understanding.
\textbf{SciFigCue} improves structured reasoning without model updates by making fine-grained connection cues explicit, further highlighting reliable connection-structure reasoning as an important direction for future scientific VLMs.

\clearpage
\section*{Limitations}

SciFigBench provides a diagnostic benchmark for evaluating fine-grained VLM understanding of scientific method figures, but several limitations remain.
First, the current dataset is limited to AI conference papers published in 2024--2025 and therefore does not cover the diverse visual conventions, layouts, notations, and arrow semantics found in other scientific domains.
We view AI method figures as a practical first step toward broader scientific method-figure understanding; future versions will expand SciFigBench to a wider range of scientific disciplines and adapt SciFigCue to domain-specific visual conventions and arrow semantics.
Second, SciFigBench relies on an automated annotation pipeline to generate candidate content and structural annotations, and candidate quality affects annotation cost and scalability.
Given the scale of the benchmark and the complexity of scientific method figures, residual annotation errors or ambiguous cases may remain even after human verification.
We will continue to refine SciFigBench in future releases based on community feedback.
Finally, SciFigCue relies on the same automated parsing pipeline to generate test-time visual cues, making its effectiveness sensitive to upstream structural parsing quality.
Further advances in scientific figure parsing may therefore improve SciFigCue performance.
\section*{Acknowledgments}
This research was supported by the Guangdong Basic and Applied Basic Research Foundation (2026A1515011435), the Fund of the National Natural Science Foundation of China (Grant No. 62576177), Shenzhen Science and Technology Program (QNXMB20250701090801002, JCYJ20250604184027034), the Fundamental Research Funds for the Central Universities 070-63263247, Youth Interdisciplinary Research Project of Nankai University (Grant No. NKQNJC202620) and the Tianjin Key Laboratory of Visual Computing and Intelligent Perception (VCIP). Computation is supported by the Supercomputing Center of Nankai University (NKSC).
This work was also supported by the Academy for Advanced Interdisciplinary Studies, Nankai University (AAIS, NKU).

\bibliography{custom}

\appendix
\section{QA Construction Details}
\label{sec:appendix:case_stats}

This section provides additional details on SciFigBench QA construction.
We summarize the case design and statistics for each task.

\paragraph{Dense Content Perception.}
Dense Content Perception is constructed as a unified flow-path OCR query, where the model reads $K$ highlighted text regions in information-flow order. 
When $K=1$, the task reduces to single-region OCR; when $K\geq2$, it evaluates ordered reading across multiple method components. 
We retain a substantial fraction of $K=1$ examples because localized text reading provides the foundation for ordered reading across multiple regions.
Table~\ref{tab:appendix_ocr_cases} summarizes the distribution over sequence length $K$.

\begin{table}[h]
\centering
\footnotesize
\setlength{\tabcolsep}{3pt}
\begin{tabular}{lrrrrrrr}
\toprule
$K$ & 1 & 2 & 3 & 4 & 5 & 6 & Total \\
\midrule
Count & 4{,}352 & 730 & 1{,}014 & 811 & 623 & 837 & 8{,}367 \\
\bottomrule
\end{tabular}
\caption{\textbf{Dense Content Perception case distribution.}
Examples are grouped by the number of highlighted text regions $K$. $K=1$ denotes the single-region special case, while $K\geq2$ denotes flow-path multi-region reading.}
\label{tab:appendix_ocr_cases}
\end{table}

\paragraph{Edge-Level Verification.}
Edge-Level Verification balances positive and negative examples and includes hard negatives.
Reverse cases select two nodes connected by a true edge, but the true edge direction is opposite to the red-to-blue query direction.
Two-hop cases select node pairs that are connected through a two-hop path but do not have a direct edge.
Sibling-source cases select node pairs that share the same source, and sibling-target cases select pairs that share the same target.
Spatial-close cases select node pairs that are spatially close but not directly connected in the graph.
Together, these hard cases prevent models from relying on simple spatial-proximity or connectivity priors, and instead require them to judge the directionality and
directness of visible connections.
Table~\ref{tab:appendix_edge_cases} reports the resulting case distribution.

\begin{table}[t]
\centering
\small
\begin{tabular}{lr}
\toprule
Case & Count \\
\midrule
Positive & 4{,}341 \\
Negative / easy & 2{,}152 \\
Negative / reverse & 493 \\
Negative / sibling source & 317 \\
Negative / sibling target & 397 \\
Negative / spatial close & 523 \\
Negative / two-hop & 442 \\
\midrule
Total & 8{,}665 \\
\bottomrule
\end{tabular}
\caption{\textbf{Edge-Level Verification case distribution.} Positive examples contain a directed edge from the red-boxed node to the blue-boxed node. Negative examples are
grouped by construction type, including easy negatives and hard negatives.}
\label{tab:appendix_edge_cases}
\end{table}

\paragraph{Node-Level Degree Counting.}
Node-Level Degree Counting queries the in- or out-degree of a target node.
We count edges rather than distinct neighboring nodes because connections in method figures usually express specific information transfer or module dependency.
Parallel edges and self-loops are counted according to the task definition to preserve fine-grained local connectivity.
Examples are stratified by in/out query direction. The 0 bucket indicates no edge in the queried direction, although edges may exist in the opposite direction.
Sampling emphasizes the more common 1-, 2-, and 3-degree cases in method figures, while retaining 4+ cases to evaluate denser local connectivity.
This sampling design covers absent, common low-degree, and denser local connectivity patterns, evaluating whether models can both detect the absence of directional edges
and aggregate existing connections.
Table~\ref{tab:appendix_degree_cases} gives the distribution by query direction and target count bucket.

\begin{table}[t]
\centering
\footnotesize
\setlength{\tabcolsep}{3pt}
\begin{tabular}{lrrrrrr}
\toprule
Direction & 0 & 1 & 2 & 3 & 4+ & Total \\
\midrule
In & 746 & 1{,}141 & 1{,}152 & 635 & 279 & 3{,}953 \\
Out & 717 & 1{,}335 & 1{,}053 & 556 & 234 & 3{,}895 \\
\midrule
Total & 1{,}463 & 2{,}476 & 2{,}205 & 1{,}191 & 513 & 7{,}848 \\
\bottomrule
\end{tabular}
\caption{\textbf{Node-Level Degree Counting case distribution.} Examples are grouped by query direction and edge-count bucket, where 0 denotes no edge in the queried
direction and 4+ denotes denser local connectivity.}
\label{tab:appendix_degree_cases}
\end{table}

\paragraph{Path-Level Reachability.}
Path-Level Reachability asks whether a directed path exists from a highlighted source node to a highlighted target node.
Positive examples preferentially select node pairs with a shortest directed path of at least two edges, reducing overlap with direct-edge verification.
Length-one direct-edge positives are retained as fallbacks only when a paper contains no positive path of length at least two. 
Consequently, 4{,}074 of the 4{,}236 positive examples require multi-hop reasoning, with only 162 retained as length-one fallbacks.
Easy negatives select node pairs that are unreachable in the queried direction and do not fall into a specific hard-negative type.
Reverse-path negatives select node pairs where the target node can reach the source node but the source node cannot reach the target node.
Sibling-source negatives select node pairs that share the same source, and sibling-target negatives select pairs that share the same target.
Spatial-close negatives select node pairs that are spatially close but not reachable in the queried direction.
Together, these cases require models to distinguish true directed reachability from reverse reachability, shared local connectivity, and visual proximity.
Table~\ref{tab:appendix_path_cases} reports the resulting case distribution.

\begin{table}[t]
\centering
\small
\begin{tabular}{lr}
\toprule
Case & Count \\
\midrule
Positive / length=1 fallback & 162 \\
Positive / length=2 & 1{,}782 \\
Positive / length=3 & 1{,}173 \\
Positive / length=4 & 710 \\
Positive / length=5+ & 409 \\
Negative / easy & 1{,}753 \\
Negative / reverse path & 773 \\
Negative / sibling source & 329 \\
Negative / sibling target & 459 \\
Negative / spatial close & 594 \\
\midrule
Total & 8{,}144 \\
\bottomrule
\end{tabular}
\caption{\textbf{Path-Level Reachability case distribution.}
Positive examples are grouped by shortest directed path length, and negative examples are grouped by construction type.
Length-one positives are used only as fallbacks for papers without any multi-hop positive path.}
\label{tab:appendix_path_cases}
\end{table}

\paragraph{Information-Flow Tracing.}
Information-Flow Tracing asks the model to list the text labels of direct incoming or outgoing neighbors of a named target node.
To avoid ambiguous references, we select only target nodes with unique textual labels within the figure.
Each eligible target yields a pair of incoming and outgoing queries, testing the same local structure in both information-flow directions.
We group targets as in-only, out-only, or both according to whether they have incoming neighbors, outgoing neighbors, or both.
For in-only and out-only targets, the missing direction yields an empty answer, testing whether models can recognize the absence of direct neighbors.
Table~\ref{tab:appendix_content_cases} lists the resulting query-level distribution by target type, query direction, and answer type.

\begin{table}[t]
\centering
\footnotesize
\setlength{\tabcolsep}{3pt}
\begin{tabular}{lllr}
\toprule
Target type & Query direction & Answer type & Count \\
\midrule
Both & Incoming & Non-empty & 2{,}485 \\
Both & Outgoing & Non-empty & 2{,}485 \\
In only & Incoming & Non-empty & 669 \\
In only & Outgoing & Empty & 669 \\
Out only & Incoming & Empty & 669 \\
Out only & Outgoing & Non-empty & 669 \\
\midrule
\multicolumn{3}{r}{Total} & 7{,}646 \\
\bottomrule
\end{tabular}
\caption{\textbf{Information-Flow Tracing case distribution.} Counts are reported at the query level after each eligible target is expanded into one incoming and one outgoing query. Both-type targets produce two non-empty queries, whereas in-only and out-only targets produce one empty query in the missing direction.}
\label{tab:appendix_content_cases}
\end{table}

\section{Automatic Annotation Quality}
\label{sec:appendix:annotation_verification}

We evaluate automatic annotation quality over all 4{,}352 figures in SciFigBench by comparing pre-verification outputs with human-verified annotations.
OCR and node boxes are matched at IoU $\geq 0.5$, while arrow endpoints are matched within a 15-pixel distance threshold.
Primary correctness additionally requires exact normalized text agreement for OCR and the correct head/tail type for arrow endpoints.

\paragraph{Overall annotation quality.}

Table~\ref{tab:appendix:annotation_quality} reports primary precision, recall, and F1 for OCR, node segmentation, and arrow endpoint annotation.
The automatic annotations achieve an overall primary F1 of 82.21\%. OCR is relatively reliable, while structural annotations remain substantially more challenging.

\begin{table}[t]

\centering
\small
\setlength{\tabcolsep}{5pt}

\begin{tabular}{lccc}

\toprule
Component & Precision & Recall & F1 \\
\midrule
OCR               & 95.46 & 95.72 & 95.59 \\
Node segmentation & 57.01 & 71.80 & 63.56 \\
Arrow endpoint    & 67.78 & 87.07 & 76.23 \\
\midrule
Overall           & 76.96 & 88.23 & 82.21 \\
\bottomrule

\end{tabular}

\caption{\textbf{Quality of automatic annotations before human verification.}
Primary metrics use the matching criteria defined above, with additional semantic correctness requirements for OCR and arrow endpoints.
The automatic pipeline achieves an overall primary F1 of 82.21\%.}

\label{tab:appendix:annotation_quality}

\end{table}

\paragraph{Arrow endpoint diagnostics.}

Because arrow cues are an important component of the structural prompts used by SciFigCue, we further analyze arrow endpoint parsing errors.
As shown in Table~\ref{tab:appendix:arrow_endpoint_quality}, matched endpoints are usually assigned the correct head/tail type, whereas associating them with the correct nodes remains more challenging.
Errors in node segmentation and endpoint association can propagate to directed-relation recovery, since valid edges depend on correctly associating endpoints with their corresponding nodes.
Thus, structural parsing rather than OCR remains the main bottleneck in the automated pipeline, consistent with the observed sensitivity of SciFigCue to upstream cue quality.

\begin{table}[t]

\centering
\small
\setlength{\tabcolsep}{5pt}

\begin{tabular}{lr}

\toprule
Arrow endpoint diagnostic & Value \\
\midrule
Endpoint match precision & 73.49 \\
Endpoint match recall & 94.41 \\
Endpoint match F1 & 82.65 \\
Head/tail accuracy & 92.23 \\
Nearest-node accuracy & 71.33 \\
Mean endpoint distance & 0.58 px \\
\bottomrule

\end{tabular}

\caption{\textbf{Diagnostics of automatic arrow endpoint parsing.}
Endpoint matching uses a 15-pixel distance threshold. Head/tail and nearest-node accuracy evaluate directional labeling and endpoint-to-node association on matched endpoints, respectively.
All matching and accuracy values are percentages.}

\label{tab:appendix:arrow_endpoint_quality}

\end{table}

\section{Implementation Details}
\label{sec:appendix:implementation_details}

The main experiments are conducted on eight NVIDIA A40 GPUs, while additional analyses and ablations are run on eight NVIDIA RTX 3090 GPUs.
For open-source models, we set the temperature to 0 and the maximum generation length to 2{,}048 tokens, with explicit chain-of-thought (CoT) or thinking modes disabled.
Proprietary models use the same maximum generation length, with additional reasoning disabled or minimized whenever supported by the corresponding API.
Specifically, GPT-5.5 uses a maximum generation length of 2{,}048 tokens with additional reasoning disabled. Temperature is omitted due to API constraints, and no explicit CoT is enabled.
GPT-5-mini also uses a maximum generation length of 2{,}048 tokens with the lowest supported reasoning setting. Temperature is omitted, and no explicit CoT is enabled.
Gemini-2.5-Flash uses a maximum generation length of 2{,}048 tokens and a temperature of 0, with thinking disabled.
Gemini-3.1-Flash-Lite uses the same maximum generation length and temperature. Since its API does not support disabling thinking entirely, we use the lowest available thinking setting.

\section{Subset Evaluation}
\label{sec:appendix:subset_evaluation}

To enable a fair comparison with proprietary models, we evaluate all open-source models on the same 2{,}000-QA subset used for GPT and Gemini models.
Table~\ref{tab:appendix:subset_results} reports the per-model results for both open- and closed-source models on this shared subset.
As summarized in Table~\ref{tab:appendix:subset_delta}, weighted-mean performance across the 13 open-source models changes by only +0.9 percentage points on average relative to the full test set, indicating that evaluation on the shared subset does not materially alter the overall comparison.

\begin{table*}[t]

\centering
\tiny

\newcommand{\taskhead}[2]{%
  \begin{tabular}[c]{@{}c@{}}
    #1\\
    #2
  \end{tabular}
}

\newcommand{\gainfont}{\fontsize{4.2}{4.6}\selectfont}
\newcommand{\posgain}[2]{#1{\gainfont\textcolor{green!50!black}{\,(#2)}}}
\newcommand{\neggain}[2]{#1{\gainfont\textcolor{red}{\,(#2)}}}

\setlength{\tabcolsep}{4pt}

\resizebox{\textwidth}{!}{%
\begin{tabular}{l c c c c c c}

\toprule

\textbf{Model}
& \textbf{Size}
& \textbf{\taskhead{Dense Content}{Perception}}
& \textbf{\taskhead{Edge-Level}{Verification}}
& \textbf{\taskhead{Node-Level}{Degree Counting}}
& \textbf{\taskhead{Path-Level}{Reachability}}
& \textbf{\taskhead{Information-Flow}{Tracing}} \\

\midrule

\textsc{RandomChoice} & -- & -- & 50.0 & -- & 50.0 & -- \\

\midrule

\rowcolor{gray!10}
\multicolumn{7}{l}{\emph{Closed-source models}} \\

Gemini-2.5-Flash
& -- & 72.7 & \underline{66.0} & 42.3 & \textbf{55.3} & 42.4 \\

Gemini-3.1-Flash-Lite
& -- & 71.2 & \textbf{67.3} & \textbf{55.8} & \underline{54.1} & \underline{46.1} \\

GPT-5-mini
& -- & 51.3 & 49.8 & 24.8 & 49.3 & 26.0 \\

GPT-5.5
& -- & \textbf{76.8} & 55.8 & 34.7 & 51.8 & 38.6 \\

\midrule

\rowcolor{gray!10}
\multicolumn{7}{l}{\emph{Open-source models}} \\

InternVL3.5-2B~\citep{internvl35}
& 2B & 35.2 & 37.7 & 31.0 & 48.1 & 20.2 \\

InternVL3.5-8B~\citep{internvl35}
& 8B & 49.4 & 44.9 & 38.0 & 41.5 & 29.1 \\

InternVL3.5-14B~\citep{internvl35}
& 14B & 53.5 & 47.2 & 35.4 & 47.0 & 29.3 \\

InternVL3.5-38B~\citep{internvl35}
& 38B & 55.3 & 51.7 & 46.0 & 41.9 & 26.2 \\

LLaVA-OV-1.5-4B~\citep{llavaonevision}
& 4B & 48.2 & 44.7 & 35.4 & 38.5 & 28.4 \\

LLaVA-OV-1.5-8B~\citep{llavaonevision}
& 8B & 56.7 & 46.0 & 32.8 & 44.2 & 28.1 \\

Qwen3-VL-4B~\citep{qwen3vl}
& 4B & 53.0 & 50.0 & 38.0 & 44.2 & 28.9 \\

Qwen3-VL-8B~\citep{qwen3vl}
& 8B & 56.4 & 50.5 & 38.0 & 46.3 & 32.8 \\

Qwen3-VL-32B~\citep{qwen3vl}
& 32B & 64.8 & 56.8 & 43.8 & 47.3 & 36.1 \\

Qwen3.5-VL-2B
& 2B & 21.8 & 45.8 & 21.9 & 48.6 & 18.2 \\

Qwen3.5-VL-4B
& 4B & 45.4 & 49.2 & 37.6 & 48.4 & 27.5 \\

Qwen3.5-VL-9B
& 9B & 58.5 & 53.8 & 39.8 & 45.6 & 38.5 \\

Qwen3.5-VL-27B
& 27B & \underline{72.8} & 59.8 & \underline{48.9} & 49.1 & \textbf{49.1} \\

\bottomrule

\end{tabular}
}%

\caption{\textbf{Results on the shared 2{,}000-QA evaluation subset.}
All values are percentages, using the same metrics as Table~\ref{tab:main_results}.
All closed- and open-source models are evaluated on the same subset.
Best and second-best results in each column are shown in \textbf{bold} and \underline{underlined}, respectively.}

\label{tab:appendix:subset_results}

\end{table*}

\begin{table}[t]

\centering
\small
\setlength{\tabcolsep}{3.5pt}

\resizebox{\columnwidth}{!}{%
\begin{tabular}{lcccccc}

\toprule
Group & $\Delta$DCP & $\Delta$Edge & $\Delta$Degree
& $\Delta$Reach. & $\Delta$IFT & $\Delta$Mean \\
\midrule

Open-source
& +1.5 & +1.2 & -0.1 & -0.2 & +2.5 & +0.9 \\

\bottomrule

\end{tabular}%
}

\caption{\textbf{Average changes from full-test to subset evaluation.}
Values are percentage-point differences computed as subset score minus full-test score.
$\Delta$Mean denotes the corresponding change in the weighted aggregate across tasks.}

\label{tab:appendix:subset_delta}

\end{table}

\section{Task-Specific Diagnostic Analysis}
\label{sec:appendix:diagnostic_analysis}

This section further analyzes model performance across task-specific subgroups to reveal fine-grained behaviors hidden by aggregate results.
The representative proprietary model, GPT-5.5, uses the same 2{,}000-QA evaluation subset, while the remaining models use the full test set.

\begin{table*}[t]

\centering
\scriptsize
\renewcommand{\arraystretch}{0.95}
\newcommand{\diagwidth}{0.75\textwidth}

\begin{tabularx}{\diagwidth}{l*{6}{Y}}
\toprule
\multicolumn{7}{l}{\rule[-3pt]{0pt}{10pt}{\fontsize{9}{10}\selectfont\textbf{(a) Dense Content Perception}}} \\
Model & $K{=}1$ & $K{=}2$ & $K{=}3$ & $K{=}4$ & $K{=}5$ & $K{=}6$ \\
\midrule

GPT-5.5
& 83.4 & 68.4 & 65.6 & 75.5 & 68.2 & 71.6 \\

Qwen3-VL-32B
& 69.8 & 44.4 & 47.8 & 54.1 & 63.7 & 66.5 \\

Qwen3.5-VL-9B
& 76.0 & 47.1 & 41.9 & 42.8 & 48.9 & 46.6 \\

\rowcolor{blue!4}
+ SciFigCue
& 84.0 & 66.7 & 56.6 & 59.0 & 64.3 & 61.1 \\

Qwen3.5-VL-27B
& 75.9 & 58.0 & 60.9 & 65.4 & 73.6 & 71.2 \\

\rowcolor{blue!8}
+ SciFigCue
& 85.2 & 79.9 & 76.5 & 79.6 & 82.3 & 80.1 \\

\bottomrule
\end{tabularx}

\vspace{5pt}

\begin{tabularx}{\diagwidth}{l*{5}{Y}}
\toprule
\multicolumn{6}{l}{\rule[-3pt]{0pt}{10pt}{\fontsize{9}{10}\selectfont\textbf{(b) Edge-Level Verification}}} \\
Model & Neg. & Pos. & Easy & Reverse & Two-Hop \\
\midrule

GPT-5.5
& 54.7 & 57.0 & 64.7 & 35.3 & 46.7 \\

Qwen3-VL-32B
& 87.9 & 20.2 & 96.1 & 94.5 & 43.7 \\

Qwen3.5-VL-9B
& 86.6 & 15.1 & 93.7 & 95.1 & 44.8 \\

\rowcolor{blue!4}
+ SciFigCue
& 64.4 & 49.0 & 73.7 & 81.5 & 15.2 \\

Qwen3.5-VL-27B
& 96.4 & 22.2 & 98.8 & 96.6 & 90.0 \\

\rowcolor{blue!8}
+ SciFigCue
& 87.9 & 52.3 & 95.4 & 88.8 & 62.4 \\

\bottomrule
\end{tabularx}

\vspace{5pt}

\begin{tabularx}{\diagwidth}{l*{5}{Y}}
\toprule
\multicolumn{6}{l}{\rule[-3pt]{0pt}{10pt}{\fontsize{9}{10}\selectfont\textbf{(c) Node-Level Degree Counting}}} \\
Model & 0 & 1 & 2 & 3 & 4+ \\
\midrule

GPT-5.5
& 36.5 & 44.2 & 26.3 & 33.3 & 22.2 \\

Qwen3-VL-32B
& 50.1 & 73.5 & 33.1 & 26.3 & 9.0 \\

Qwen3.5-VL-9B
& 49.8 & 34.4 & 62.8 & 18.6 & 13.1 \\

\rowcolor{blue!4}
+ SciFigCue
& 41.2 & 47.7 & 66.1 & 14.3 & 9.9 \\

Qwen3.5-VL-27B
& 43.3 & 79.3 & 47.2 & 24.2 & 14.2 \\

\rowcolor{blue!8}
+ SciFigCue
& 44.4 & 80.6 & 42.2 & 30.0 & 14.2 \\

\bottomrule
\end{tabularx}

\vspace{5pt}

\begin{tabularx}{\diagwidth}{l*{6}{Y}}
\toprule
\multicolumn{7}{l}{\rule[-3pt]{0pt}{10pt}{\fontsize{9}{10}\selectfont\textbf{(d) Path-Level Reachability}}} \\
Model & Neg. & Pos. & $L{=}2$ & $L{=}3$ & $L{=}4$ & $L{\ge}5$ \\
\midrule

GPT-5.5
& 21.7 & 79.6 & 83.1 & 74.4 & 79.6 & 78.6 \\

Qwen3-VL-32B
& 80.2 & 20.9 & 20.0 & 22.2 & 23.5 & 20.5 \\

Qwen3.5-VL-9B
& 74.9 & 19.9 & 18.2 & 21.2 & 23.8 & 19.6 \\

\rowcolor{blue!4}
+ SciFigCue
& 37.8 & 65.0 & 62.2 & 66.4 & 68.3 & 71.6 \\

Qwen3.5-VL-27B
& 77.0 & 25.5 & 23.5 & 27.8 & 28.6 & 27.9 \\

\rowcolor{blue!8}
+ SciFigCue
& 62.7 & 49.7 & 48.5 & 51.7 & 52.0 & 49.1 \\

\bottomrule
\end{tabularx}

\vspace{5pt}

\begin{tabularx}{\diagwidth}{l*{4}{Y}}
\toprule
\multicolumn{5}{l}{\rule[-3pt]{0pt}{10pt}{\fontsize{9}{10}\selectfont\textbf{(e) Information-Flow Tracing}}} \\
Model & 0 & 1 & 2--3 & 4+ \\
\midrule

GPT-5.5
& 12.8 & 43.8 & 48.4 & 20.0 \\

Qwen3-VL-32B
& 34.3 & 34.8 & 39.0 & 33.2 \\

Qwen3.5-VL-9B
& 68.0 & 24.3 & 30.1 & 23.0 \\

\rowcolor{blue!4}
+ SciFigCue
& 57.4 & 31.1 & 34.0 & 24.2 \\

Qwen3.5-VL-27B
& 42.2 & 43.9 & 44.9 & 37.0 \\

\rowcolor{blue!8}
+ SciFigCue
& 21.4 & 47.0 & 45.8 & 36.9 \\

\bottomrule
\end{tabularx}

\caption{\textbf{Task-specific diagnostic analysis.}
All values are percentages.
\textbf{(a)} DCP is grouped by sequence length $K$.
\textbf{(b)} Edge reports answer labels and representative negative subtypes.
\textbf{(c)} Degree is grouped by the degree count in the queried direction.
\textbf{(d)} Reach. reports answer labels and shortest-path length $L$ for positive examples; rare length-one fallbacks are omitted from the $L$ breakdown.
\textbf{(e)} IFT is grouped by the number of ground-truth neighbors.
GPT-5.5 uses the 2{,}000-QA subset, while the remaining models use the full test set.
These breakdowns reveal distributed-reading difficulty, direct/indirect relation confusion, dense-connectivity failures, and answer biases.}

\label{tab:appendix:diagnostic_breakdowns}

\end{table*}

\paragraph{Dense Content Perception.}

We break down Dense Content Perception by the number of queried regions $K$ to examine the transition from localized reading to ordered reading across multiple regions.
As shown in panel (a) of Table~\ref{tab:appendix:diagnostic_breakdowns}, most models exhibit a clear performance gap between $K=1$ and $K\ge2$.
SciFigCue consistently improves Qwen3.5-VL across all $K\ge2$ settings, further indicating that explicit information-flow cues facilitate ordered reading across distributed regions.

\paragraph{Edge-Level Verification.}

Panel (b) of Table~\ref{tab:appendix:diagnostic_breakdowns} breaks down Edge-Level Verification by answer label and representative negative subtypes.
Several models exhibit substantial positive--negative performance imbalance, indicating that aggregate accuracy can obscure answer bias.
Two-hop negatives further expose confusion between direct and indirect relations; for example, Qwen3.5-VL-9B drops from 93.7 on easy negatives to 44.8 on two-hop negatives.

\paragraph{Node-Level Degree Counting.}

Panel (c) of Table~\ref{tab:appendix:diagnostic_breakdowns} stratifies Node-Level Degree Counting by the degree count in the queried direction.
Performance degrades substantially as local connectivity becomes denser.
For example, Qwen3.5-VL-27B scores 79.3 at degree 1 but only 14.2 at degree 4+, revealing dense-connectivity aggregation as a major bottleneck.

\paragraph{Path-Level Reachability.}

Panel (d) of Table~\ref{tab:appendix:diagnostic_breakdowns} analyzes Path-Level Reachability by answer label and shortest-path length for positive examples.
The results reveal substantial answer bias rather than a simple monotonic trend with path length.
For example, Qwen3.5-VL-27B scores 77.0 and 25.5 on negative and positive examples, respectively, while SciFigCue changes these scores to 62.7 and 49.7.
This contrast shows that aggregate reachability accuracy can conceal strong answer-class preferences.

\paragraph{Information-Flow Tracing.}

Panel (e) of Table~\ref{tab:appendix:diagnostic_breakdowns} stratifies Information-Flow Tracing by ground-truth neighbor count to examine empty-answer bias.
Several models show large discrepancies between zero-neighbor and non-empty cases, revealing empty-answer bias hidden by aggregate scores.
For example, Qwen3.5-VL-9B scores 68.0 on zero-neighbor cases but only 24.3 on one-neighbor cases.
SciFigCue raises the one-neighbor score to 31.1 and substantially narrows the gap between zero- and one-neighbor cases.

\section{Color Robustness}
\label{sec:appendix:color_robustness}

Because query-box colors may interact with colors already present in method figures, we further evaluate the robustness of SciFigBench to different box-color schemes.
Prior work has shown that visual-prompt color can affect VLM behavior~\citep{shtedritski2023does}. We therefore test three perceptually distinct schemes: the original red/blue, green/purple, and cyan/orange palettes.
The evaluation covers three VLMs and the full QA sets of the three box-grounded structural tasks: Edge-Level Verification, Node-Level Degree Counting, and Path-Level Reachability.
Although absolute scores vary across color schemes, the weighted-score ranking remains unchanged in all settings: Qwen3.5-VL-9B $>$ InternVL3.5-14B $>$ LLaVA-OV-1.5-8B.
Thus, changing query-box colors does not alter the model-level comparison, suggesting that the model-level conclusions are not tied to the original red/blue palette.

\begin{table*}[t]

\centering
\small
\setlength{\tabcolsep}{5pt}

\begin{tabular}{llccccc}

\toprule
Model & Scheme & Edge & Degree & Reach. & Weighted & $\Delta$Weighted \\
\midrule

Qwen3.5-VL-9B
& Red/blue
& 50.8 & 41.4 & 46.3 & 46.3 & +0.0 \\

& Green/purple
& 54.7 & 40.3 & 52.3 & 49.3 & +3.0 \\

& Cyan/orange
& 50.8 & 36.5 & 49.4 & 45.8 & -0.5 \\

\midrule

InternVL3.5-14B
& Red/blue
& 47.7 & 35.7 & 46.3 & 43.4 & +0.0 \\

& Green/purple
& 51.5 & 35.3 & 49.2 & 45.6 & +2.2 \\

& Cyan/orange
& 48.4 & 34.5 & 47.4 & 43.7 & +0.3 \\

\midrule

LLaVA-OV-1.5-8B
& Red/blue
& 45.4 & 34.5 & 40.2 & 40.2 & +0.0 \\

& Green/purple
& 48.7 & 34.2 & 45.0 & 42.9 & +2.7 \\

& Cyan/orange
& 48.3 & 33.0 & 44.5 & 42.2 & +2.0 \\

\bottomrule

\end{tabular}

\caption{\textbf{Robustness to query-box color schemes.}
All values are percentages.
Weighted denotes the task-size-weighted aggregate across Edge, Degree, and Reach., and $\Delta$Weighted is the absolute change relative to the original red/blue scheme.}

\label{tab:appendix:color_robustness}

\end{table*}

\paragraph{Local color contrast.}

We further examine whether query boxes are difficult to distinguish from local figure backgrounds.
We use CIELAB $\Delta E_{76}$ to measure contrast between box boundaries and figure backgrounds, treating boxes with a median boundary contrast below 20 as low-contrast.
As shown in Table~\ref{tab:appendix:color_visibility}, low-contrast cases are negligible across all three schemes, indicating that query boxes do not systematically blend into figure backgrounds.
Moreover, color contrast and model performance are not monotonic: green/purple has lower median contrast than red/blue but yields higher weighted scores for all three models.
We therefore do not claim color invariance; rather, these results indicate that the main conclusions are not driven by systematic color interference or a single red/blue convention.

\begin{table*}[t]

\centering
\small
\setlength{\tabcolsep}{3.5pt}

\resizebox{0.55\textwidth}{!}{%
\begin{tabular}{lrrrrr}

\toprule
Scheme
& Boxes
& Low-contrast
& Fraction
& Median $\Delta E$
& 10th pct. $\Delta E$ \\
\midrule

Red/blue
& 41{,}466
& 0
& 0.0000\%
& 113.37
& 107.54 \\

Green/purple
& 41{,}466
& 0
& 0.0000\%
& 96.76
& 96.17 \\

Cyan/orange
& 41{,}466
& 4
& 0.0096\%
& 59.17
& 56.00 \\

\bottomrule

\end{tabular}%
}

\caption{\textbf{Local visibility of query boxes under different color schemes.}
Larger $\Delta E_{76}$ indicates stronger local box/background contrast.
The 10th percentile summarizes lower-visibility cases.}

\label{tab:appendix:color_visibility}

\end{table*}

\section{Dataset Annotation Examples}
\label{sec:appendix:annotation_examples}

This section presents representative annotation examples from SciFigBench to make the annotation format more concrete.
Each example shows both the content annotation and the structural annotation, corresponding to textual content and directed structural relations in the figure.
The content annotation localizes text regions, supporting Dense Content Perception and QA construction that requires textual evidence.
The structural annotation marks component regions, arrow endpoints, and directed connections between components, supporting relation verification, degree counting, reachability, and information-flow tracing.
In the visualizations, green arrow endpoints indicate arrow tails, while red arrow endpoints indicate arrow heads.
These examples illustrate the dense text, free-form layouts, and arrow-based connections common in real method figures, as well as how SciFigBench QA tasks are grounded in fine-grained annotations.

\section{Task Prompts}
\label{sec:appendix:prompts}

We use fixed task prompts for all examples within each task type, except that Information-Flow Tracing replaces the \texttt{\{target\}} placeholder with the target node text.
The figures below show the prompts in the form used for evaluation.

\section{Visual Analysis}
\label{sec:appendix:visual_analysis}

This section complements the aggregate evaluation with representative visual cases, illustrating typical model errors across all five tasks and the remaining failure modes of SciFigCue.
These cases make the specific failure sources hidden by aggregate scores more explicit and further illustrate which errors can be mitigated by explicit structural cues and which remain unresolved.

\paragraph{Task-level qualitative cases.}

We present a representative failure case for each of the five tasks in a separate figure so that fine-grained textual and structural details remain legible.
The selected cases include incorrect ordering in multi-region text reading, misinterpretation of arrow direction in a reverse-edge case, confusion between incoming and outgoing degree, failure to compose multi-hop directed relations, and incorrect information-flow tracing after locating the target node.

\paragraph{SciFigCue failure modes.}

\begin{figure*}[t]
\centering
\includegraphics[width=0.92\textwidth]{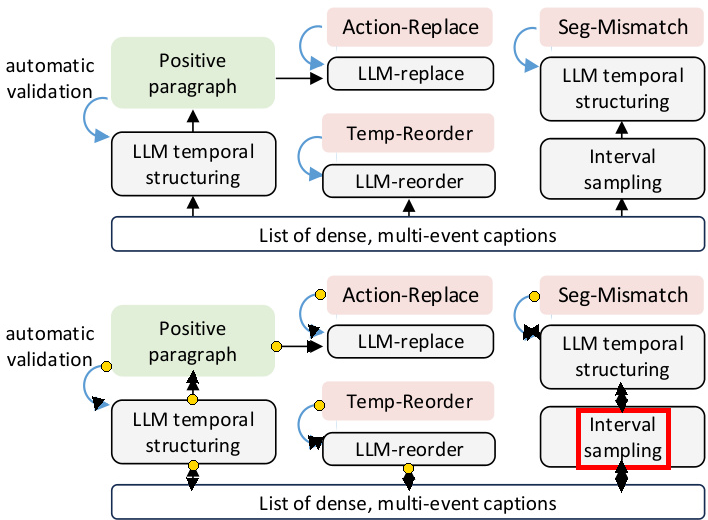}
\caption{\textbf{Parser-induced cue error in Information-Flow Tracing for Qwen3.5-VL-9B.}
The canonical input is shown above and the SciFigCue input is shown below.
The model answers correctly under the canonical input, but fails under SciFigCue because incorrectly parsed arrow cues are overlaid near the selected node and obscure valid arrow evidence in the original figure.
\textbf{Ground truth:} \texttt{\detokenize{["LLM temporal structuring"]}}.
\textbf{Canonical answer:} \texttt{\detokenize{["LLM temporal structuring"]}}.
\textbf{SciFigCue answer:} \texttt{\detokenize{["List of dense, multi-event captions"]}}.}
\label{fig:scifigcue_failure_parser_cue}
\end{figure*}

\begin{figure*}[t]
\centering
\includegraphics[width=0.92\textwidth]{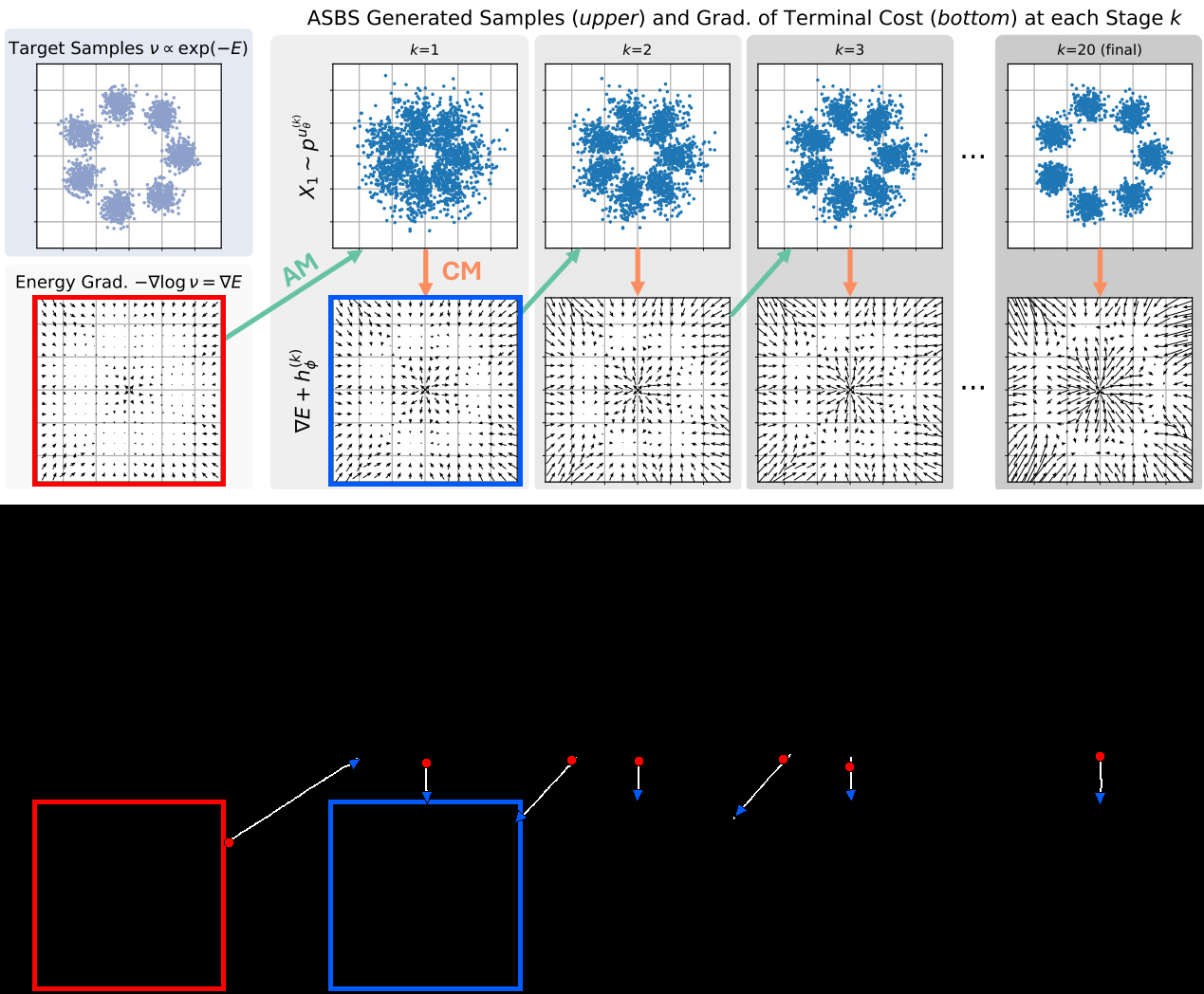}
\caption{\textbf{Residual reasoning failure in Edge-Level Verification for Qwen3.5-VL-9B.}
The canonical input is shown above, and SciFigCue augments it with the additional visual input shown below; both images are provided to the model.
This example is a negative two-hop case asking whether a direct edge exists from the red-boxed node to the blue-boxed node.
SciFigCue provides correct structural cues, but the two nodes are connected only through an intermediate node rather than by a direct edge.
The model nevertheless answers \texttt{"true"} instead of the ground-truth \texttt{"false"}, indicating persistent confusion between direct and indirect relations despite correct cues.}
\label{fig:scifigcue_failure_cue_induced}
\end{figure*}

Although SciFigCue improves multiple structure-dependent tasks, its effectiveness remains limited by upstream cue fidelity and the model's residual structural reasoning ability.
Inspection of unsuccessful cases reveals two major sources of failure: parser-induced cue errors and residual reasoning errors despite correct structural cues.
Parser-induced errors, including missed or spurious arrows, incorrect endpoint attachment, and erroneous head/tail labels, can produce incomplete or misleading structural cues.
In dense regions, incorrect overlays may also obscure valid arrow evidence, amplifying upstream parsing errors.
Figure~\ref{fig:scifigcue_failure_parser_cue} illustrates this failure: the model answers the canonical input correctly, but misparsed overlaid cues lead to an incorrect SciFigCue prediction.
A second source of failure is residual model reasoning: even when the structural cues are correct, the model may still miscount connections, confuse direct and indirect relations, or fail to compose multi-hop paths.
Figure~\ref{fig:scifigcue_failure_cue_induced} shows such a two-hop hard case with correct cues, where the model still mistakes an indirect connection for a direct edge.
These cases align with the cue-quality sensitivity analysis, showing that SciFigCue remains limited by both automatic parsing errors and residual VLM reasoning failures.

\FloatBarrier

\begin{figure*}[t]
\centering
\includegraphics[width=0.92\textwidth,trim=0pt 0pt 0pt 0pt,clip]{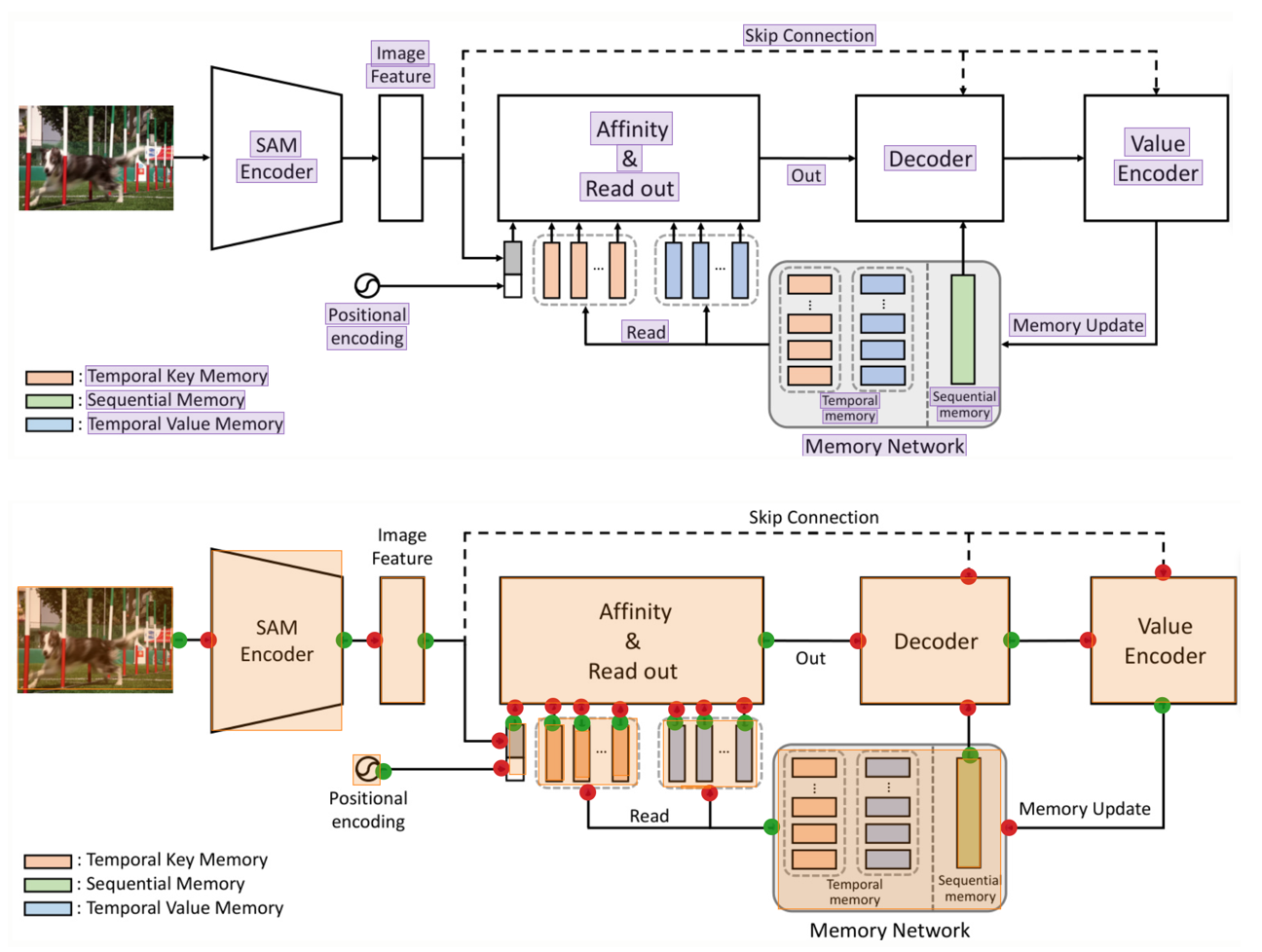}
\caption{\textbf{Dataset annotation example 1.}
Content and structural annotations for a real method figure.}
\label{fig:annotation_example_1}
\end{figure*}

\begin{figure*}[t]
\centering
\includegraphics[width=0.92\textwidth,trim=0pt 0pt 0pt 0pt,clip]{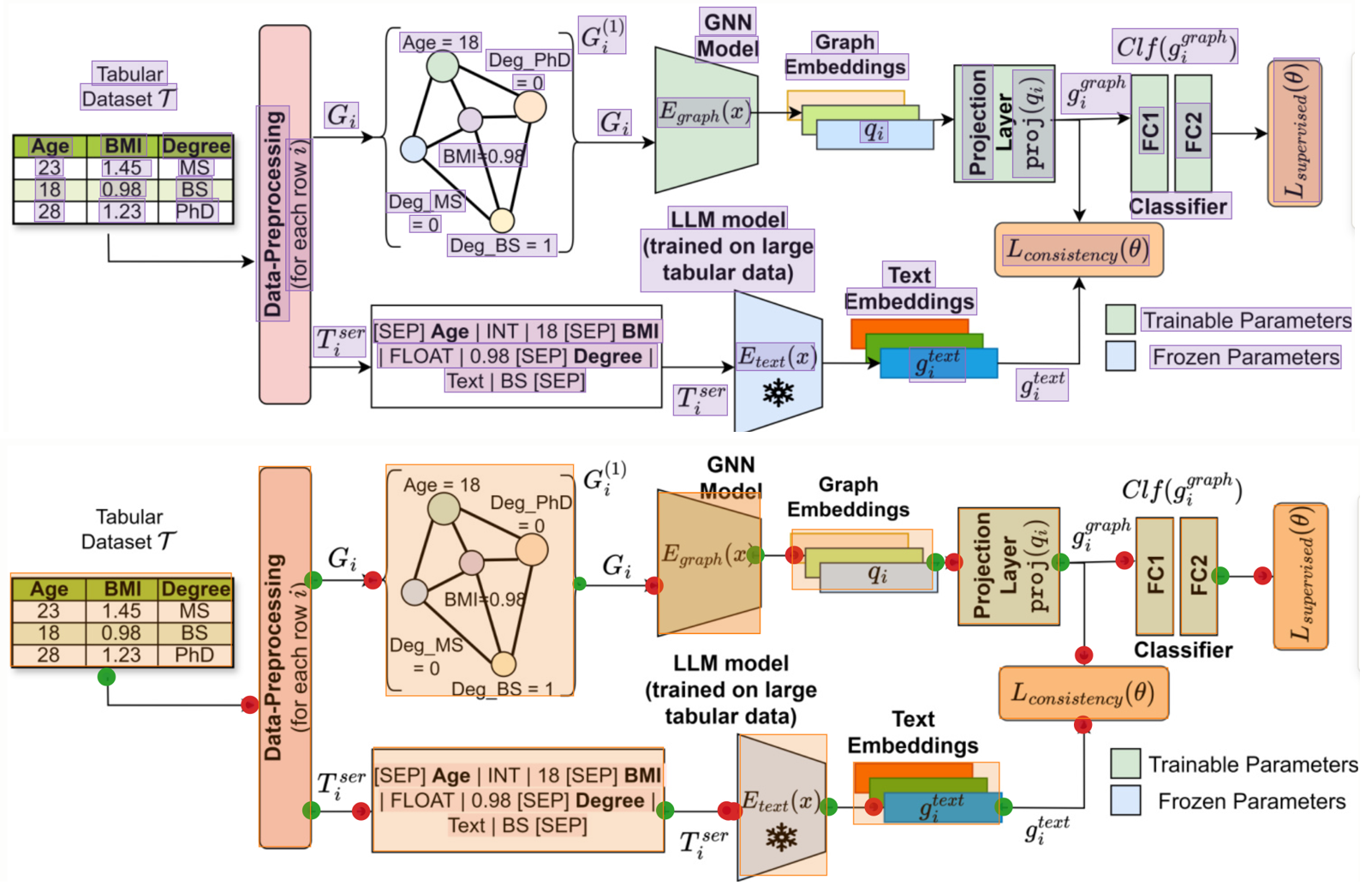}
\caption{\textbf{Dataset annotation example 2.}
Content and structural annotations for a real method figure.}
\label{fig:annotation_example_2}
\end{figure*}

\begin{figure*}[t]
\centering
\includegraphics[width=0.92\textwidth,trim=0pt 0pt 0pt 0pt,clip]{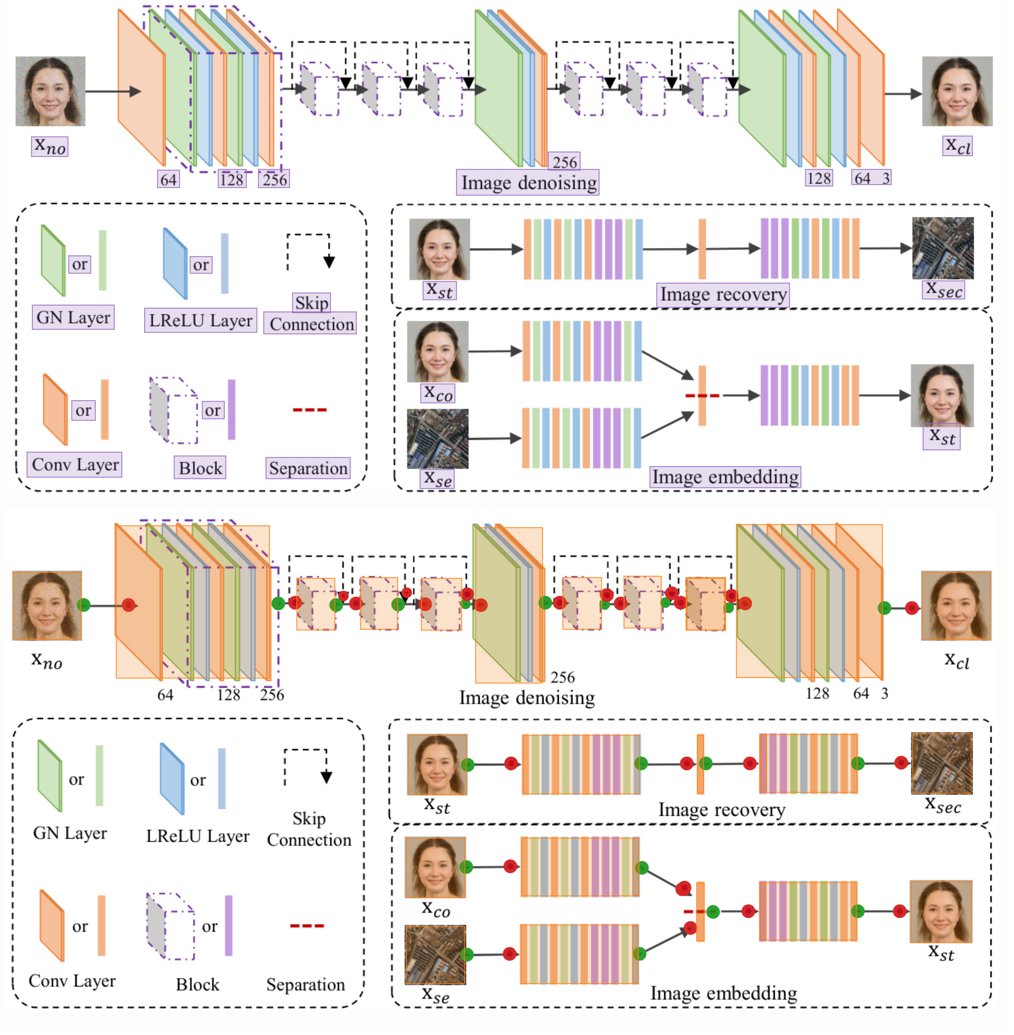}
\caption{\textbf{Dataset annotation example 3.}
Content and structural annotations for a real method figure.}
\label{fig:annotation_example_3}
\end{figure*}

\begin{figure*}[t]
\centering
\includegraphics[width=0.92\textwidth,trim=75pt 20pt 75pt 20pt,clip]{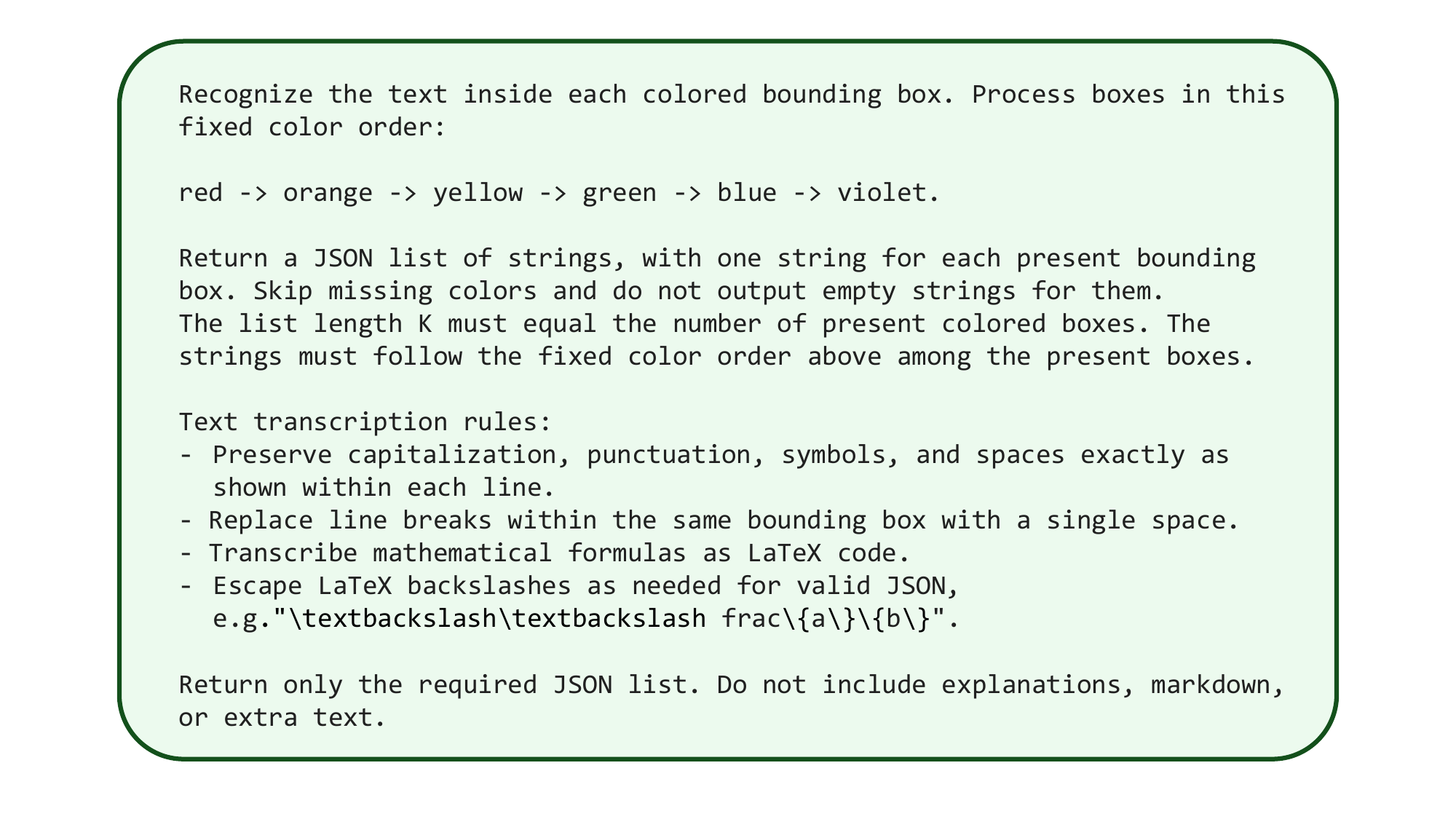}
\caption{\textbf{Prompt for Dense Content Perception.}
The prompt asks the model to read highlighted text regions in a fixed color order and return the transcriptions as a JSON list.}
\label{fig:prompt_dense_content}
\end{figure*}

\begin{figure*}[t]
\centering
\includegraphics[width=0.92\textwidth,trim=65pt 20pt 65pt 20pt,clip]{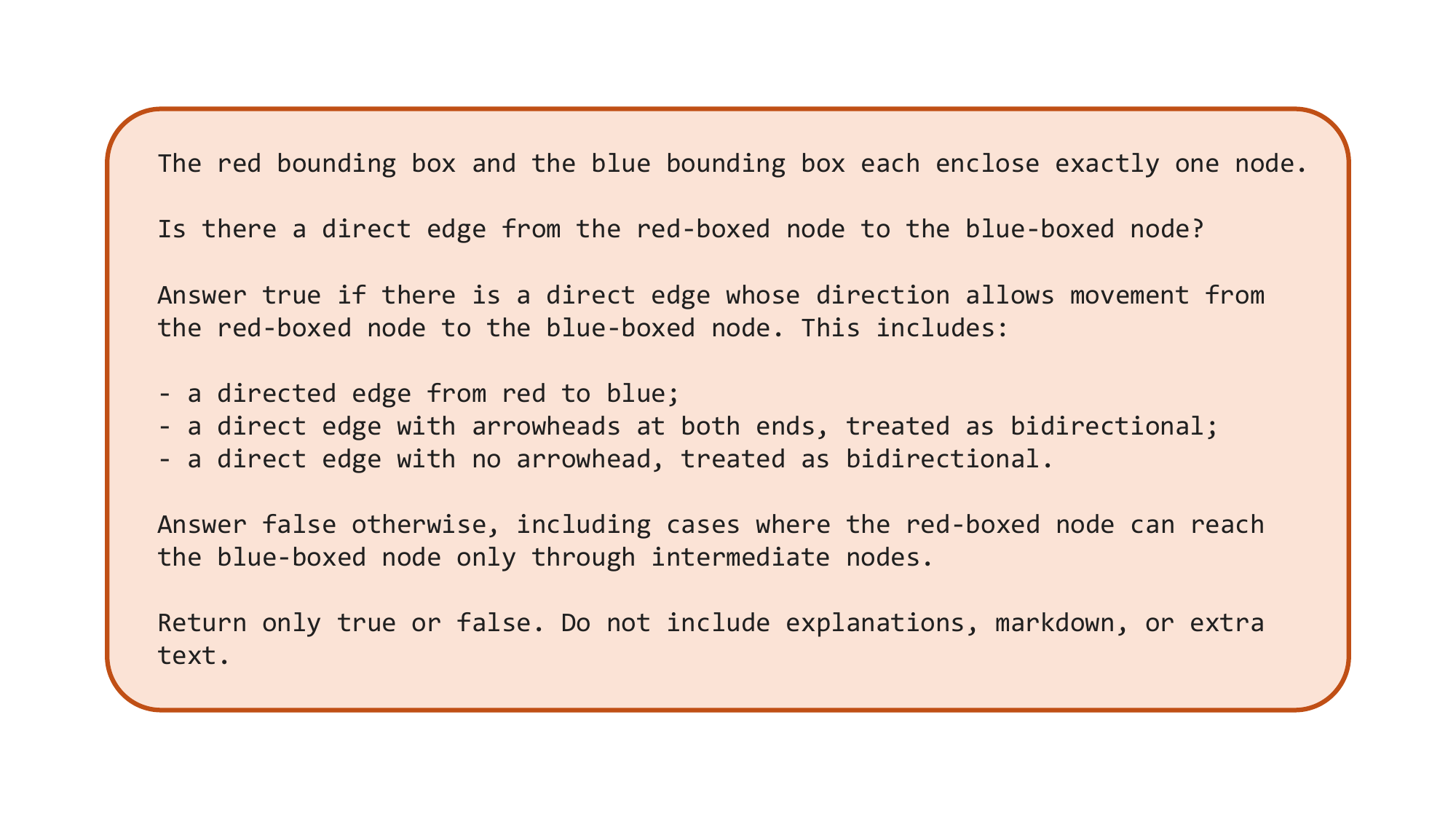}
\caption{\textbf{Prompt for Edge-Level Verification.}
The prompt asks whether a directed edge exists from the red-boxed node to the blue-boxed node under the specified edge-direction rules.}
\label{fig:prompt_edge_level}
\end{figure*}

\begin{figure*}[t]
\centering
\includegraphics[width=0.92\textwidth,trim=65pt 20pt 65pt 20pt,clip]{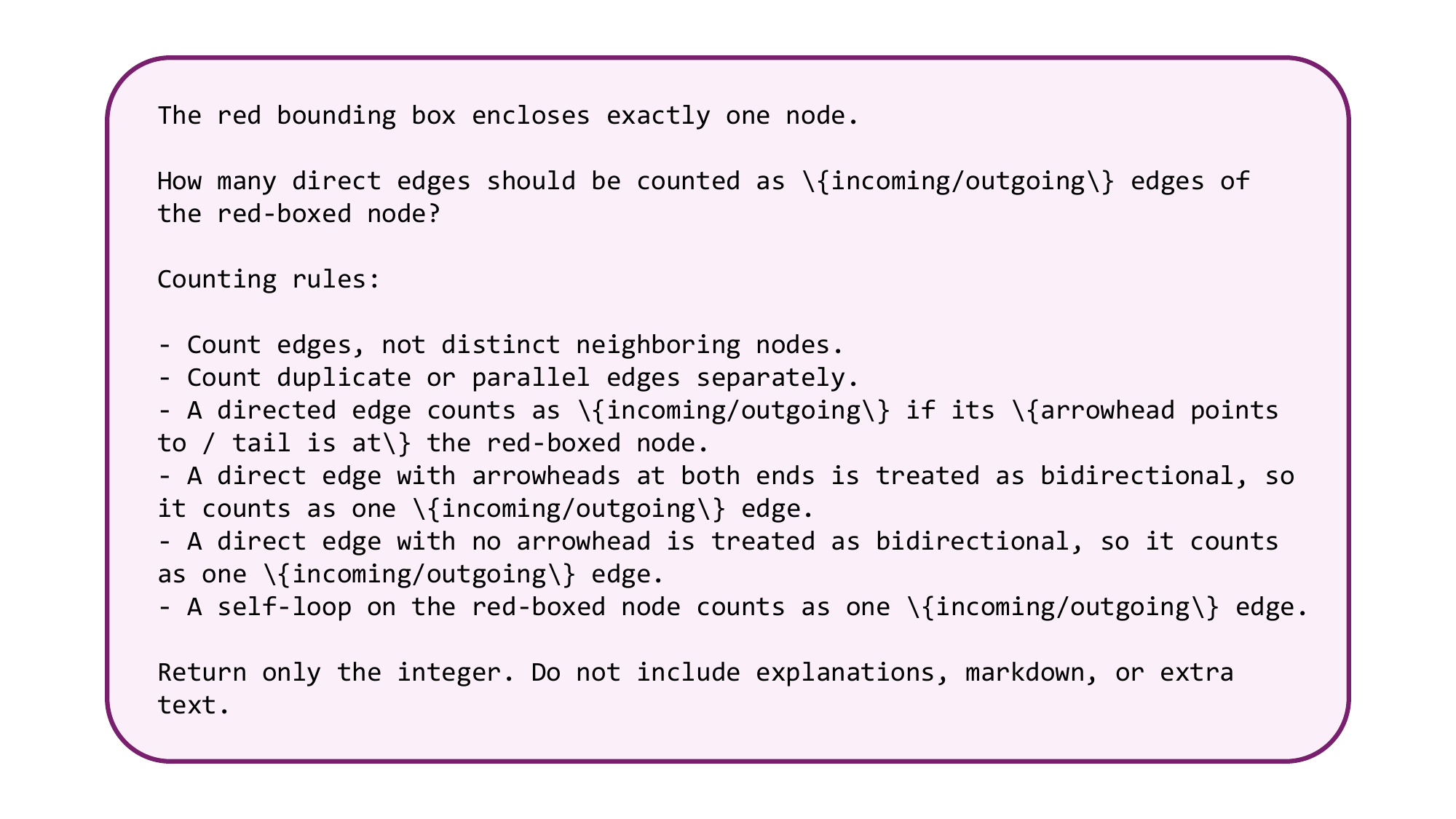}
\caption{\textbf{Prompt for Node-Level Degree Counting.}
The prompt asks the model to count incoming or outgoing edges of the red-boxed node under the specified degree-counting rules.}
\label{fig:prompt_node_level}
\end{figure*}

\begin{figure*}[t]
\centering
\includegraphics[width=0.92\textwidth,trim=65pt 20pt 65pt 20pt,clip]{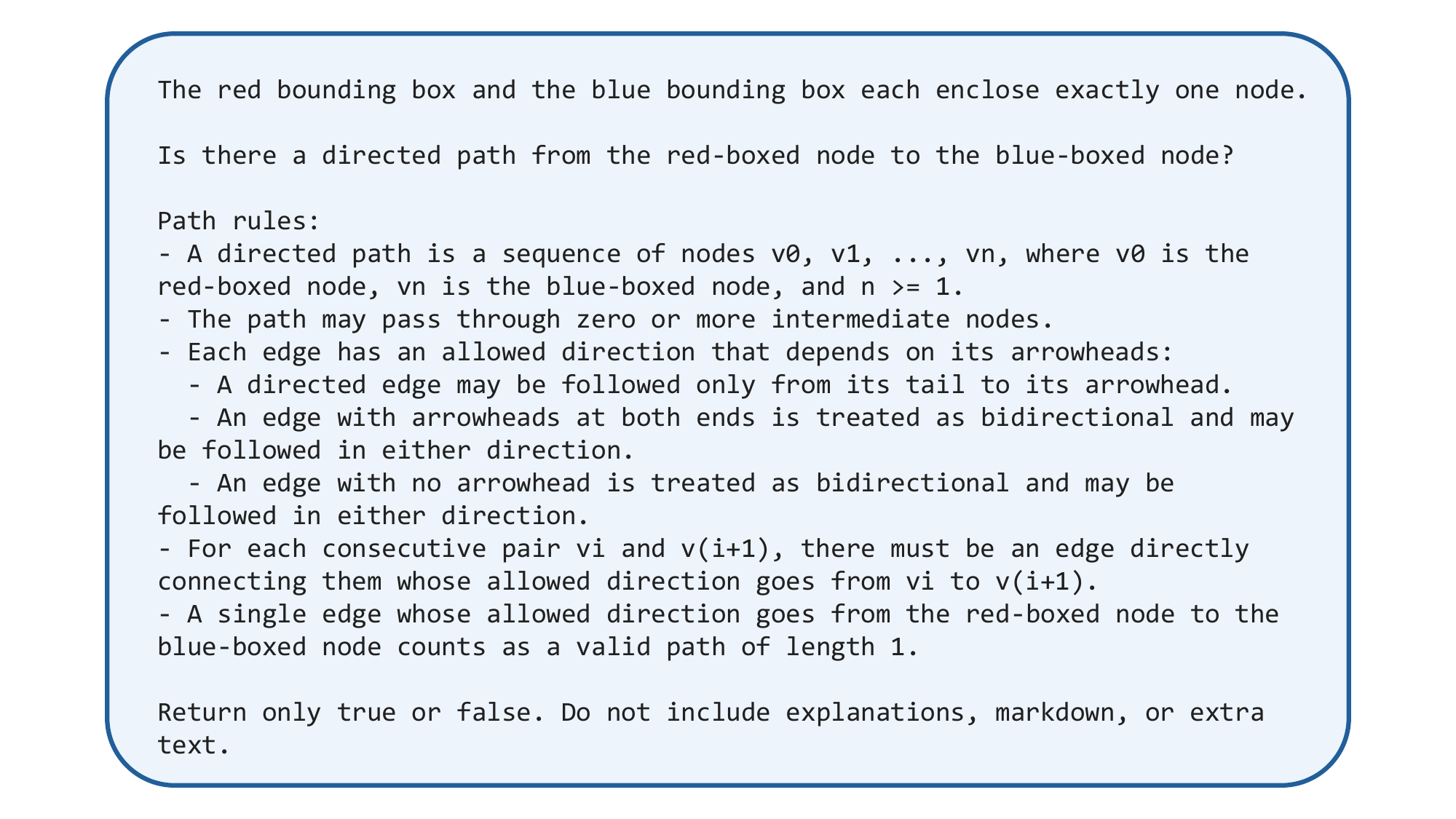}
\caption{\textbf{Prompt for Path-Level Reachability.}
The prompt asks whether a directed path exists from the red-boxed node to the blue-boxed node under the specified path-following rules.}
\label{fig:prompt_path_level}
\end{figure*}

\begin{figure*}[t]
\centering
\includegraphics[width=0.92\textwidth,trim=65pt 20pt 65pt 20pt,clip]{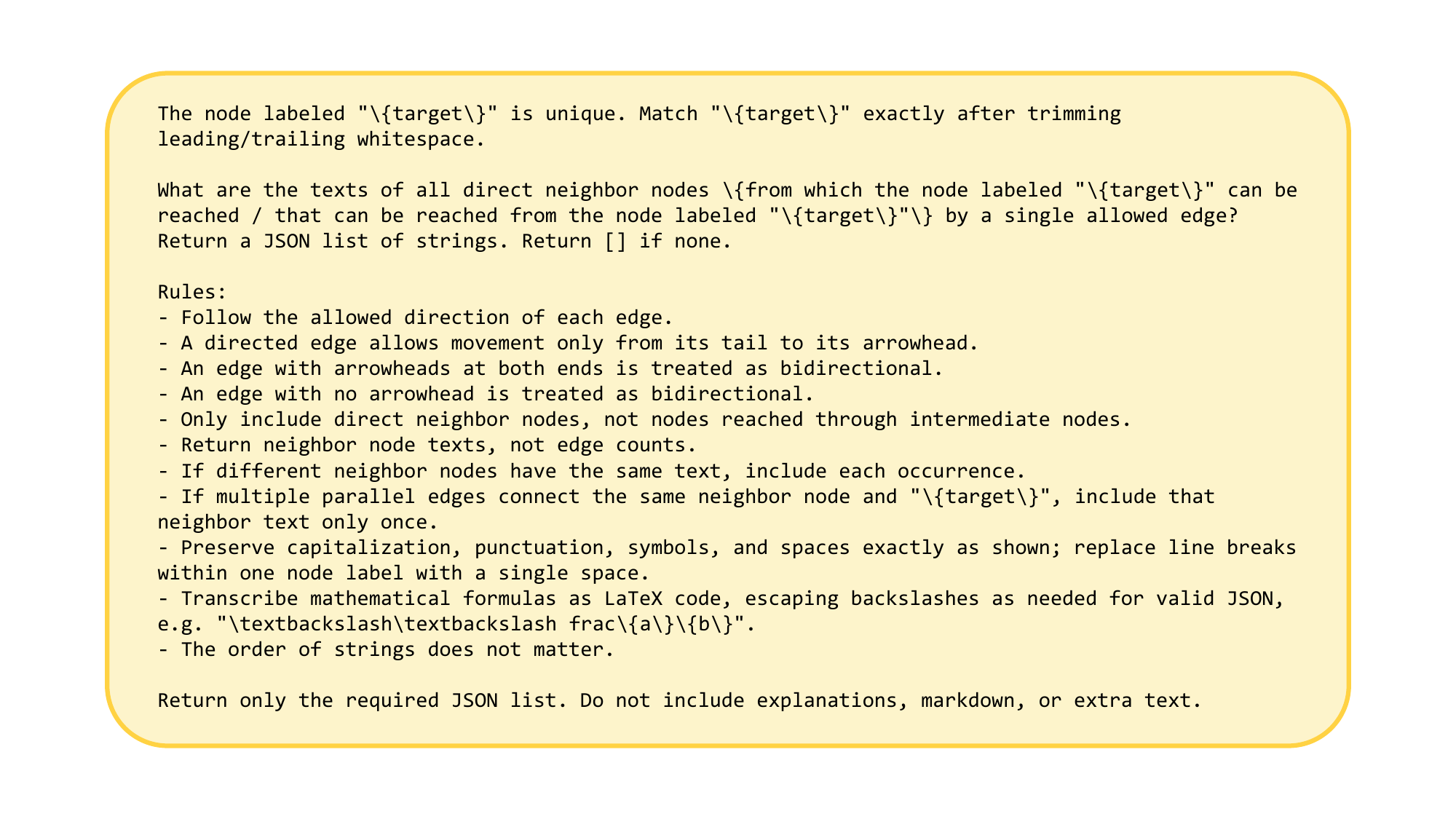}
\caption{\textbf{Prompt for Information-Flow Tracing.}
The prompt asks the model to list the text of direct incoming or outgoing neighbors for a uniquely named target node.}
\label{fig:prompt_info_flow}
\end{figure*}

\begin{figure*}[t]
\centering
\includegraphics[width=0.92\textwidth]{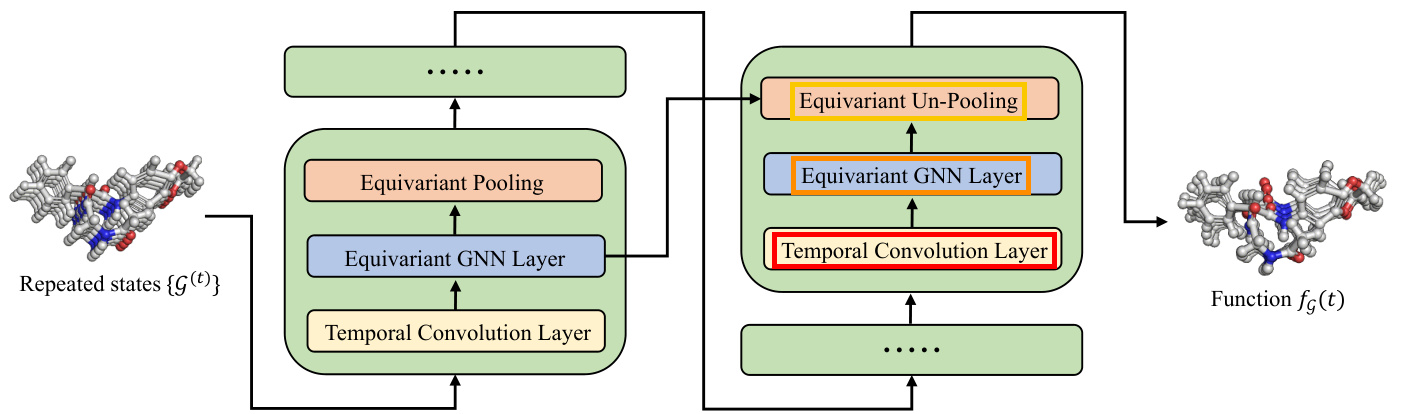}
\caption{\textbf{Dense Content Perception failure for Qwen3.5-VL-27B.}
The query asks the model to read the highlighted regions in the specified color order.
The model correctly recognizes all three text contents but swaps the order of the latter two regions, returning an incorrectly ordered sequence.
\textbf{Ground truth:} \texttt{\detokenize{["Temporal Convolution Layer", "Equivariant GNN Layer", "Equivariant Un-Pooling"]}}.
\textbf{Model answer:} \texttt{\detokenize{["Temporal Convolution Layer", "Equivariant Un-Pooling", "Equivariant GNN Layer"]}}.}
\label{fig:visual_dense_content_failure}
\end{figure*}

\begin{figure*}[t]
\centering
\includegraphics[width=0.92\textwidth]{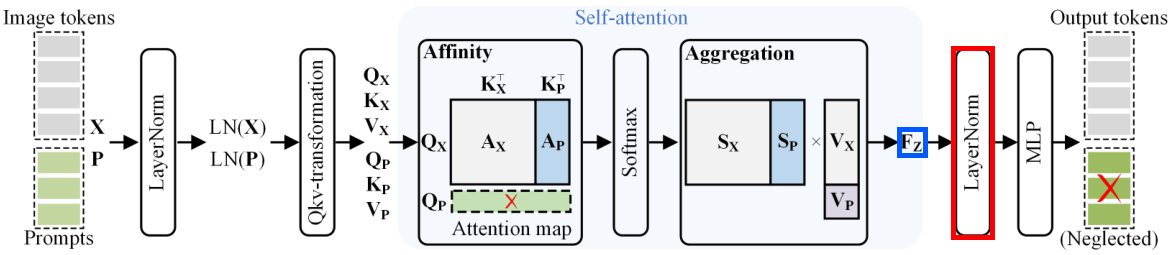}
\caption{\textbf{Edge-Level Verification failure for InternVL3.5-8B.}
The query asks whether a directed edge exists from the red-boxed node to the blue-boxed node.
The figure instead contains the reverse blue-to-red connection, but the model answers \texttt{"true"} rather than the ground-truth \texttt{"false"}, indicating a failure to correctly follow arrow direction.}
\label{fig:visual_edge_failure}
\end{figure*}

\begin{figure*}[t]
\centering
\includegraphics[width=0.92\textwidth]{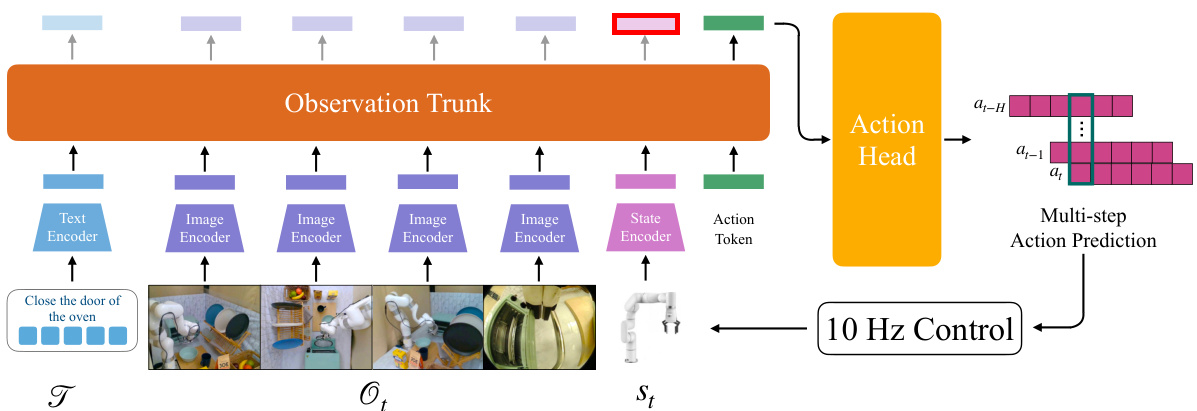}
\caption{\textbf{Node-Level Degree Counting failure for Qwen3.5-VL-9B.}
The query asks for the out-degree of the red-boxed node.
The node has one incoming edge but no outgoing edge, yet the model answers \texttt{1} instead of the ground-truth \texttt{0}, indicating confusion between incoming and outgoing connections.}
\label{fig:visual_degree_failure}
\end{figure*}

\begin{figure*}[t]
\centering
\includegraphics[width=0.92\textwidth]{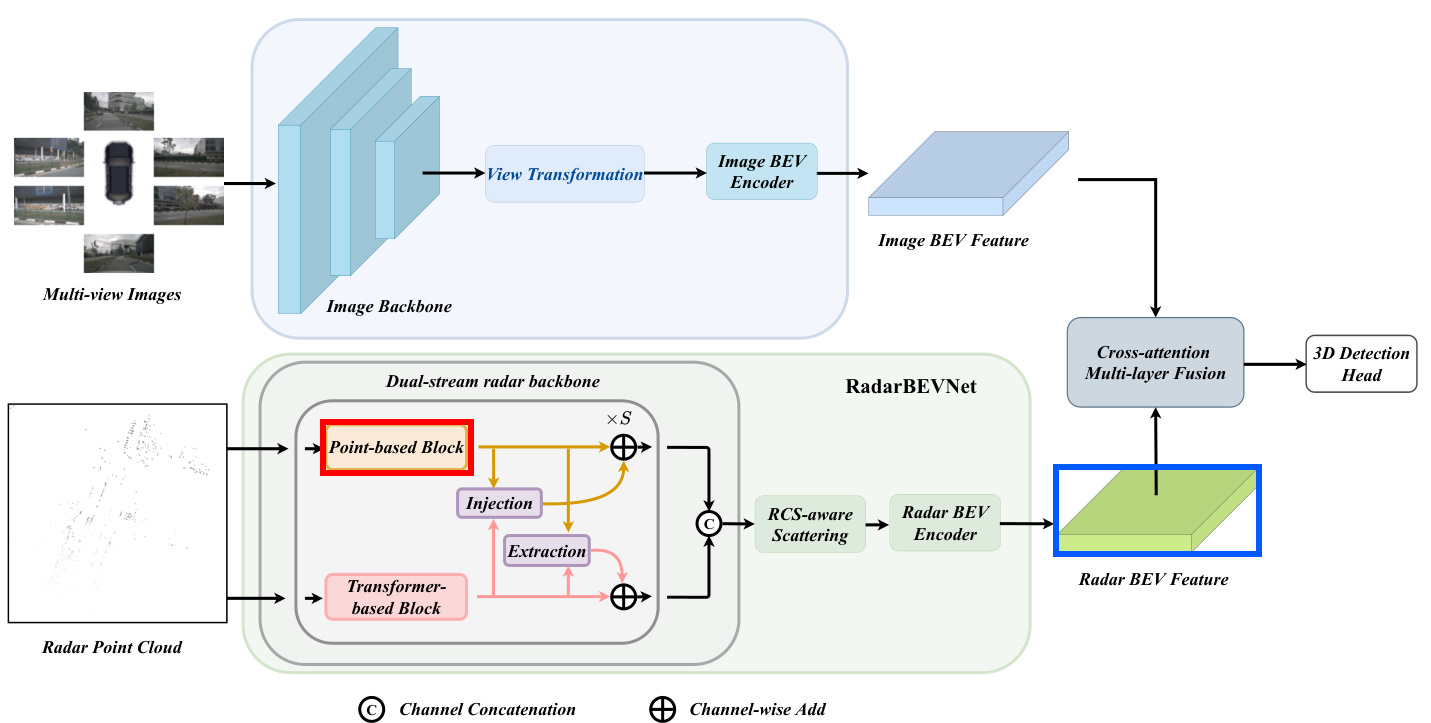}
\caption{\textbf{Path-Level Reachability failure for Qwen3.5-VL-9B.}
The query asks whether a directed path exists from the red-boxed node to the blue-boxed node.
A multi-hop directed path connects the two nodes through intermediate components, yet the model answers \texttt{"false"} instead of the ground-truth \texttt{"true"}, indicating a failure to compose multiple directed relations.}
\label{fig:visual_reachability_failure}
\end{figure*}

\begin{figure*}[t]
\centering
\includegraphics[width=0.92\textwidth]{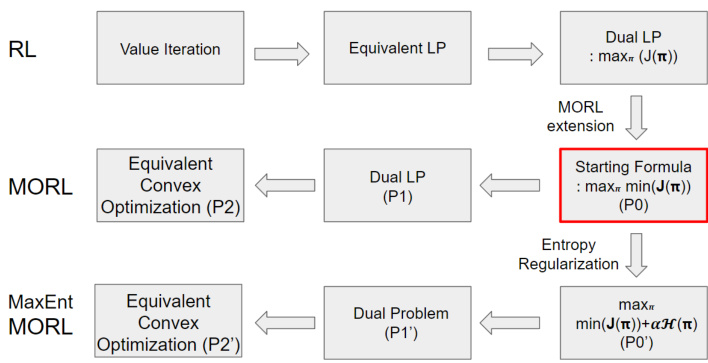}
\caption{\textbf{Information-Flow Tracing failure for Qwen3.5-VL-9B.}
The query asks for the direct neighbor reachable from the target node \textit{Value Iteration} along the specified information-flow direction.
The figure contains a direct edge from \textit{Value Iteration} to \textit{Equivalent LP}, yet the model returns an empty list instead of the ground-truth \texttt{\detokenize{["Equivalent LP"]}}, indicating a failure to trace the outgoing information flow correctly.}
\label{fig:visual_information_flow_failure}
\end{figure*}

\end{document}